\documentclass[a4paper,11pt,onecolumn]{onethree}
\usepackage{onethree-en}
\usepackage{longtable}
\graphicspath{{images/}}
\newcommand{\cmark}{\textcolor{green!60!black}{\ding{51}}}
\newcommand{\xmark}{\textcolor{red!70!black}{\ding{55}}}

\makeatletter

\newcolumntype{"}{@{\hskip\tabcolsep\vrule width 1pt\hskip\tabcolsep}}
\makeatother
\title{MatOS: An Agent Operating System for Materials Research}

\author{AI Materials Team}
\affiliation{Zhongguancun Academy\,\,$\cdot$\,\,Zhongguancun Institute of Artificial Intelligence, Beijing, China}
\contribution{{\sffamily Team e-mail:} \email{aimaterials@bza.edu.cn}}

\date{October 2026}
\checkdata[Code]{\url{https://github.com/DingyangLyu/MatOS}}

\abstract{Materials research relies on data, knowledge, and simulation tools that are scattered across many sources and continue to grow. A single study often involves multi-day workflows that iterate between literature analysis, computation, and experiment. Large language model (LLM) agents have made rapid progress in this setting: they can retrieve materials data, operate simulation codes, and, when coupled with autonomous laboratories, close design--synthesis--characterization loops. However, in materials research, extending these agents with new capabilities, keeping long-running tasks on track, and enabling them to evolve by learning from past tasks and user feedback remain difficult. Here we present MatOS, an agent operating system for materials research built on three core components: (1) a layered scientific asset graph that organizes data, knowledge, tools, skills, roles, and workflows as reusable assets that agents can combine and extend for new tasks; (2) a stateful adaptive workflow that runs research tasks as explicit execution graphs with scientific checks, so that long-running tasks can be resumed and traced; and (3) evidence-driven asset evolution, which revises these assets from execution records and user feedback, so that experience gained in one task can be reused in later ones. On ResearchClawBench, MatOS improves on the previously best-performing agent by 11.15\% with the same base model and extends the cost--score Pareto frontier. With a common base model, MatOS also achieves higher success rates than Codex CLI and Claude Code, exceeding Codex CLI by 2.13\% on MatTools and by 4.17\% on SciAgentGym. In eight research cases, MatOS narrowed more than 30 million candidate structures to 97 solid-electrolyte candidates, identified 89 halide variants with higher computed conductivity than their parent structures, and linked the force errors of 11 pretrained interatomic potentials to specific local atomic environments. It also supported the development of two new methods: a training-data selection strategy that reduces interatomic-potential errors by 3.1\% relative to DIRECT at the same data budget, and an alloy melting-temperature model with 13.6\% lower error than the best baseline. Together, these results show that organizing research capabilities as reusable, evolving assets enables agents to carry out long, multistage materials studies reliably, from large-scale screening to the development of new methods.
}
\begin{document}
\maketitle
\newpage
\tableofcontents
\thispagestyle{plain}
\setcounter{page}{1}
\newpage
\section{Introduction}

Advanced materials underpin progress in clean energy, manufacturing, and national security\cite{nrc2008icme,nstc2011mgi}. Developing them brings together simulation, synthesis, and characterization\cite{nrc2008icme,nstc2011mgi,depablo2014mgi}, guided by knowledge distributed across publications, databases, and laboratory records\cite{tshitoyan2019unsupervised,olivetti2020datadriven}. Agents based on large language models (LLMs)\cite{achiam2023gpt4,deepseekai2025r1,yang2025qwen3,openai2025gpt5,deepseekai2025v32,moonshot2025kimik2} can now plan multistep tasks and call external tools\cite{yao2023react,anthropic2025claudecode,openai2025codex,google2025geminicli}, while autonomous laboratories automate a growing share of experimental work\cite{szymanski2023alab,burger2020robotic,macleod2020selfdriving,zhang2025crest}. In chemistry and materials science, such agents have been applied to autonomous synthesis, automated simulation, materials discovery, and other tasks\cite{bran2024chemcrow,boiko2023coscientist,song2025chemagents,mcnaughton2024cactus,chiang2025llamp,zhang2024honeycomb,chaudhari2026matsciagent,kang2024chatmof,liu2025vaspilot,lv2026chatmat,zhang2026matclaw,wang2025matagent,ghafarollahi2026sparksmatter,shi2026matbrain}.

However, building dependable research infrastructure from these capabilities is still difficult\cite{alvizmeza2026copilots}, because materials research imposes three specific requirements on an agent system. First, the resources of the field keep expanding: new databases, simulation codes, and machine-learning potentials appear continually\cite{jain2013materialsproject,barrosoluque2026omat24,batatia2025macemp}, so the system must be able to take on new capabilities without being rebuilt. Second, a single study spans literature analysis, computation, and experiment and often runs for days or longer. Each stage builds on earlier results, so errors propagate from one stage to the next\cite{nstc2011mgi,chiang2025llamp}; the system must therefore keep long-running tasks on track and its computed results reproducible\cite{huber2020aiida,lejaeghere2016repro}. Third, much of the knowledge needed to run a study well, such as suitable simulation timescales and sampling strategies, is gained through experience and rarely written down\cite{zhang2026matclaw}. Every research cycle adds to this experience\cite{depablo2014mgi,szymanski2023alab}, so the system should retain it and reuse it in later tasks\cite{gao2026selfevolving,du2026agentopt}.

Existing work has made progress on each of these requirements, but three common limitations remain (Section~\ref{sec:related-work}). First, new capabilities are mostly added as tools. Materials and chemistry agents connect language models to expert-designed tools, tool hubs, MCP servers, or tool knowledge graphs\cite{bran2024chemcrow,zhang2024honeycomb,liu2025vaspilot,anthropic2024mcp,ding2025scitoolagent}, yet specialized multi-agent pipelines remain the most common design\cite{alvizmeza2026copilots}, so adding a new code or experimental stage often requires new tool functions and changes to the pipeline\cite{zhang2026matclaw}. Second, recovery mechanisms such as workflow restarts, agent checkpoints, and the automatic restart of failed VASP calculations operate on single jobs or saved states\cite{jain2015fireworks,huber2020aiida,langchain2026langgraphpersistence,liu2025vaspilot}. When an agent revises its plan during a study, the system must also determine which earlier results remain valid, while repeated runs of the same agent can still produce different outcomes\cite{rabanser2026reliability}. Third, methods for experience learning retain reflections, insights, workflows, skills, or tool knowledge\cite{shinn2023reflexion,zhao2024expel,wang2025awm,wang2024voyager,tang2026scitoolagentevo,gao2026selfevolving,memory2026agents}, and most were developed on general-domain tasks. Experience in materials research, such as simulation protocols, failure diagnoses, and validation criteria, must be checked before reuse because a flawed protocol can invalidate later results; turning it into checked, reusable capabilities remains an open problem.

To meet these requirements, we present \emph{MatOS}, an agent operating system for materials research. Just as an operating system manages programs, files, and processes for the applications running on it, MatOS manages research capabilities, task execution, and accumulated experience for the agents that use them (Figure~\ref{fig:matos-overview}). Its basic unit is the \emph{scientific asset}: data, knowledge, tools, skills, roles, and workflows are each registered with an identifier, declared dependencies, and provenance. A layered scientific asset graph links these assets through typed relations, so a new database, code, or method can be added by registering assets rather than rebuilding the agent. Each role is an agent defined by a responsibility prompt and its own subgraph of linked assets. Given a research task in natural language, a role composes the relevant data, knowledge, skills, tools, and workflows from its subgraph, and a stateful adaptive workflow carries out the task as an explicit execution graph with scientific checks. The workflow records graph snapshots, node states, and artifacts, so that long-running tasks can be paused, resumed, and traced; changes to the graph, proposed by the agent or a researcher, are admitted only after structural, policy, and compilation checks. After each task, evidence-driven asset evolution critiques the execution traces and user feedback, then revises the assets responsible for a problem. Each revision is adopted only after structural checks and semantic review, without changing foundation-model parameters.

Our main contributions are:
\begin{enumerate}
    \item We introduce the scientific asset as the basic unit for organizing research capabilities in materials science. Six asset types and the typed relations among them form a layered scientific asset graph, which we use to organize 3,135 materials datasets, 65 knowledge assets, 194 tools, 386 skills, 51 roles, and 104 workflows, together with 48 integrated materials software packages (Appendix~\ref{app:asset-coverage}).
    \item We design a stateful adaptive workflow runtime that executes directed acyclic graphs and cyclic state machines with checked plan updates and persistent execution records. We pair it with evidence-driven asset evolution, which revises assets through evidence critique, focused editing, and dual acceptance, so that experience from one task improves later ones.
    \item We evaluate MatOS on three agent benchmarks and eight research cases. It improves on the previously best-performing agent\cite{mira2026general} on ResearchClawBench\cite{xu2026researchclawbench} by 11.15\% with the same base model, and outperforms Codex CLI and Claude Code on MatTools\cite{liu2025mattools} and SciAgentGym\cite{shen2026sciagentgym}. Seven cases range from screening more than 30 million candidate structures to developing new methods (Section~\ref{sec:scientific-cases}), and the eighth, a CFDBench reproduction with OpenFOAM\textsuperscript{\textregistered}\footnote{\normalsize\OpenFOAMAcknowledgement} extends the evaluation to continuum flow (Appendix~\ref{app:cfdbench-cavity}).
\end{enumerate}

\section{System Architecture}
\label{sec:methods}

\begin{figure}[t]
    \centering
    \includegraphics[width=\textwidth]{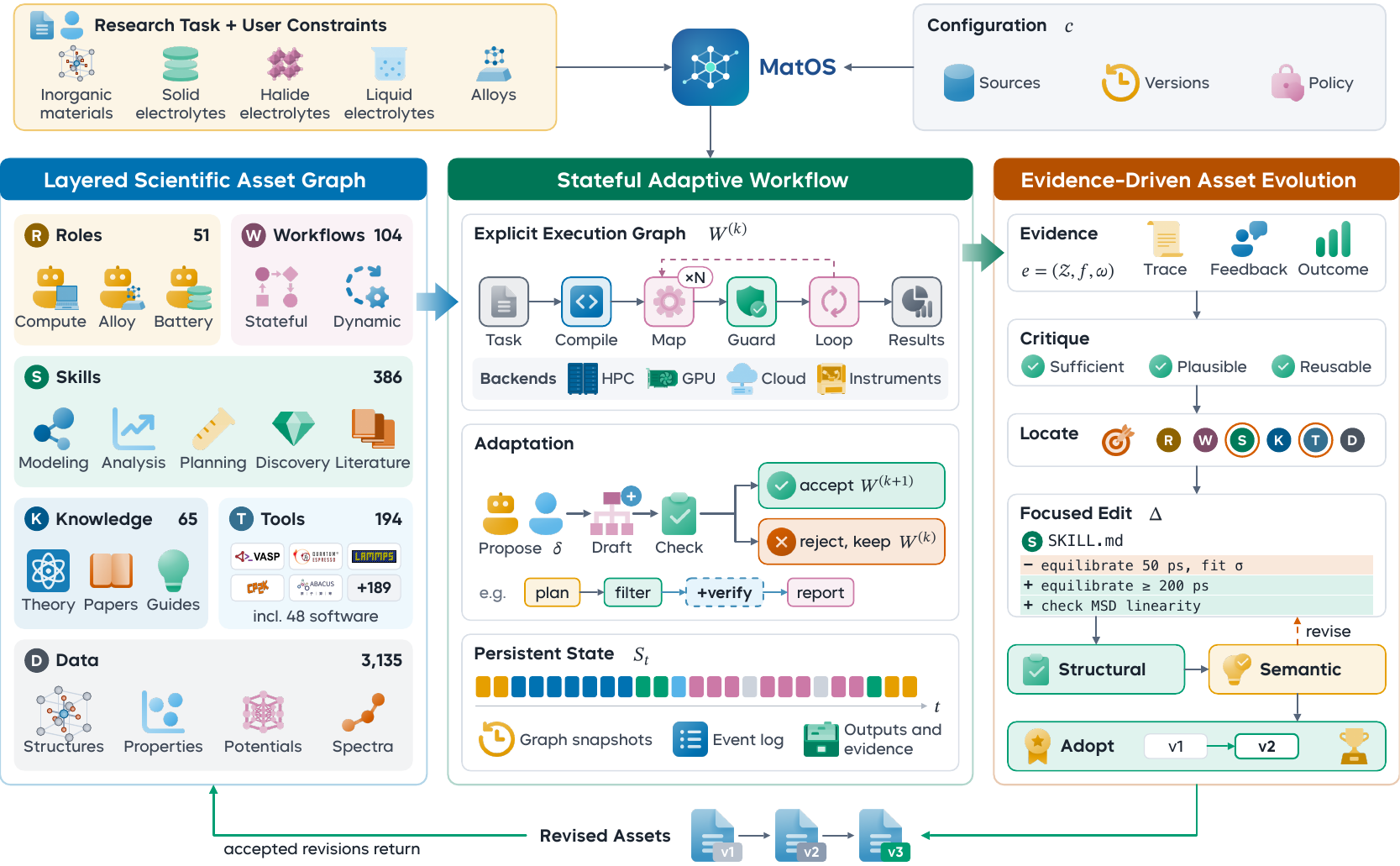}
    \caption{Overview of the MatOS architecture. Research tasks and user constraints drive the stateful adaptive workflow, while the layered scientific asset graph supplies composable research capabilities. Execution traces and user feedback lead to asset revisions through critique, focused editing, structural checks, and semantic review. Accepted revisions return to the asset graph, and benchmark validation informs subsequent updates.}
    \label{fig:matos-overview}
\end{figure}

MatOS organizes the capabilities needed for materials research as scientific assets of six types: data, knowledge, tools, skills, roles, and workflows. Three components operate on these assets (Figure~\ref{fig:matos-overview}). The \emph{layered scientific asset graph} (Section~\ref{sec:asset-graph}) defines the asset types and the relations through which assets are combined. When a task arrives, a role, which is an agent defined by a responsibility prompt and its linked assets, composes the assets the task requires, and the \emph{stateful adaptive workflow} (Section~\ref{sec:stateful-workflow}) carries out the task as an execution graph while recording its state. \emph{Evidence-driven asset evolution} (Section~\ref{sec:asset-evolution}) then uses these records and user feedback to revise the assets, so that later tasks start from improved capabilities.

\subsection{Layered Scientific Asset Graph}
\label{sec:asset-graph}

\begin{figure}[t]
    \centering
    \includegraphics[width=\textwidth]{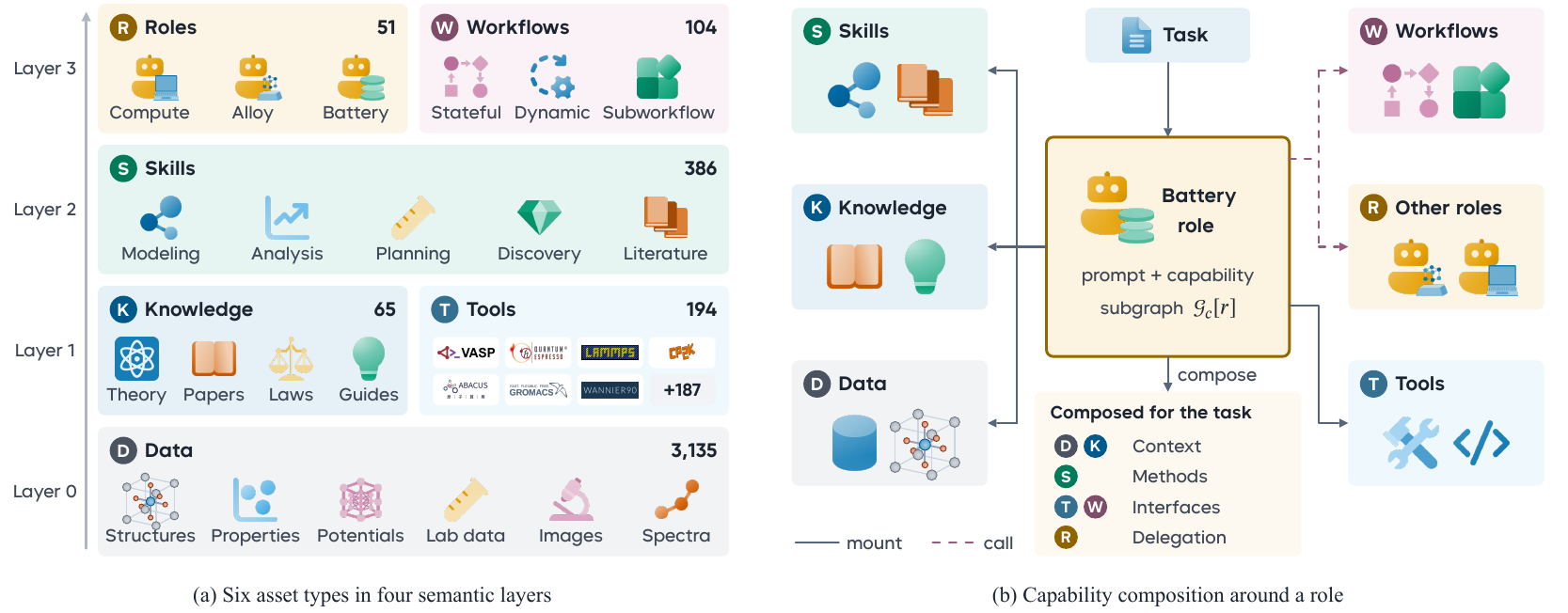}
    \caption{Organization of six scientific asset types and an example centered on a role. (a) Successive abstractions from data through knowledge, tools, and skills to roles and workflows. (b) How a role selects and combines data, knowledge, tools, skills, and workflows in an asset graph.}
    \label{fig:layered-asset-graph}
\end{figure}

\paragraph{Definition of a scientific asset.}
A scientific asset is a reusable unit of research capability. Each asset has a unique identifier, a type, and declared dependencies, and it either holds its content directly or refers to an external resource such as a database or a simulation code. Identifiers and declared dependencies make it possible to trace where an asset comes from and where it is used, following FAIR principles and materials workflow practice\cite{wilkinson2016fair,jain2015fireworks,huber2020aiida,ganose2025atomate2}. We consider six asset types, denoted $D$, $K$, $T$, $S$, $R$, and $W$ (Table~\ref{tab:scientific-assets} and Figure~\ref{fig:layered-asset-graph}).

\begin{table}[htb]
    \centering
    \small
    \caption{Symbols and scientific meanings of the six MatOS asset types.}
    \label{tab:scientific-assets}
    \begin{tabular}{@{}clp{0.70\textwidth}@{}}
        \toprule
        Symbol & Type & Scientific meaning \\
        \midrule
        $D$ & Data & Records of materials systems and their computational or experimental processes, including measurements, simulation outputs, structures, parameters, and references to data sources. \\
        $K$ & Knowledge & Generalizations from observations and their conditions of applicability, including physical laws, empirical judgments, and knowledge about the use of methods. \\
        $T$ & Tool & Callable capabilities centered on materials solvers, including software interfaces, retrieval and processing routines, validation scripts, and instrument interfaces. \\
        $S$ & Skill & A method for a class of problems that specifies how knowledge and tools are selected and used and how results are checked. \\
        $R$ & Role & An agent configuration defined by a responsibility prompt and the assets linked to it, which specify the capabilities it may select, combine, and invoke. \\
        $W$ & Workflow & A reusable process description that organizes tools, roles, and subworkflows through nodes, dependencies, and control conditions. \\
        \bottomrule
    \end{tabular}
\end{table}

In a typical materials research task, a researcher uses data $D$ and knowledge $K$ to define the question and its constraints, then follows a skill $S$ to select and organize tools $T$ for solution and validation. A role $R$ reasons and decides within the task, while a workflow $W$ coordinates multiple solution and validation stages. Results become new data and may in turn revise knowledge and methods. These dependencies among the six types determine the layers and edges of the asset graph.

\paragraph{Semantic layers.}
We assign the six types to four semantic layers according to their typical prerequisites (Figure~\ref{fig:layered-asset-graph}a). Data form the base layer, supplying the evidence from which knowledge is formed and against which results are checked. Knowledge and tools form the second layer, because selecting a tool and interpreting its output require domain knowledge. Skills combine data, knowledge, and tools into methods and form the third layer. Roles and workflows form the organizational layer: roles assemble capabilities, and workflows orchestrate processes. Formally, the asset graph is $\mathcal G=(V,E,\tau)$, where $V$ is the set of assets, $\tau$ assigns each asset one of the six types $\mathcal T=\{D,K,T,S,R,W\}$, and each edge $(a,\lambda,b)\in E$ is a directed relation of type $\lambda$ from asset $a$ to asset $b$.

\paragraph{Relation types and composition.}
Edges specify how two assets are related, and their direction follows the meaning of the relation: the source asset refers to the target. We distinguish six relation types, defined in Table~\ref{tab:asset-relations}: dependency, mount, call, containment, evidence, and verification. A pair of assets may have more than one relation; for example, a role may both mount a workflow and call it.

\begin{table}[!htb]
    \centering
    \small
    \setlength{\tabcolsep}{4pt}
    \renewcommand{\arraystretch}{1.15}
    \caption{Semantics of directed scientific asset edges. An edge $a\rightarrow b$ points from the referencing asset to its referent. The last column lists typical type pairs.}
    \label{tab:asset-relations}
    \begin{tabular}{@{}p{2.4cm}p{7.6cm}>{\raggedright\arraybackslash}p{4.5cm}@{}}
        \toprule
        Relation & Meaning of an edge from $a$ to $b$ & Typical type connection \\
        \midrule
        Dependency & The applicability or use of $a$ requires data, knowledge, or a capability supplied by $b$. & $S\rightarrow D,K,T$; $T\rightarrow K$ \\
        Mount & $b$ becomes part of the capabilities available to $a$ for selection and composition. & $R\rightarrow D,K,T,S,W$; $S\rightarrow K,T$ \\
        Call & $a$ may delegate a solution or decision to $b$ and use the returned result. & $R\rightarrow T,R,W$ \\
        Containment & A research process described by $a$ contains a stage handled by $b$, which may be a reusable subprocess. & $W\rightarrow T,R,W$ \\
        Evidence & $b$ supplies traceable evidence for a claim, method choice, or result judgment in $a$. & $K,S,W\rightarrow D,K$ \\
        Verification & $b$ specifies or performs a check of the applicability, behavior, or result of $a$. & $K,S,W\rightarrow T,S,W$ \\
        \bottomrule
    \end{tabular}
\end{table}

The first four relations describe the prerequisites and composition of research capabilities. The last two record scientific evidence and validation and accumulate as assets are used. Layers and relation types are independent: a skill may depend on data, knowledge, and tools at once, knowledge may refer to a tool that tests it, and assets on the same layer may be connected. A role can call a workflow that contains a stage handled by the same role, so the asset graph may contain cycles. The execution order and loop conditions of a task are set by its workflow definition (Section~\ref{sec:stateful-workflow}).

\paragraph{Role definition and capability composition.}
For each task, a configuration $c$ specifies which asset sources are available and the runtime policy. We write $V_c$ for the assets available under $c$ and $\mathcal G_c$ for the graph they form. A role $r$ is defined by its responsibility prompt and its capability subgraph $\mathcal G_c[r]$, which contains the assets reachable from $r$ through dependency, mount, and call relations. For a given task, the role composes the capabilities it needs from this subgraph: data and knowledge provide research context, skills specify method steps, tools and workflows serve as solution interfaces, and other roles take on delegated subtasks.

Because roles and the methods they mount are separate assets, each can be reused on its own. Multiple roles can mount the same skill, and different workflows can include the same role as a stage. A compute role, for example, may invoke a battery parameterization planning workflow, in which an agent node calls an electrochemical modeling role to draft the parameterization plan with that role's responsibilities and method context. A role can also run AiiDA work chains through skill and workflow assets, leaving the internal calculations and their provenance to AiiDA\cite{huber2020aiida}.

These relations describe how capabilities are declared to fit together. To retrieve assets and locate revision targets (Section~\ref{sec:asset-evolution}), MatOS also builds an extended view $\mathcal G_c^{\mathrm{view}}$ that adds links among related assets of the same type, together with records of their use, execution, and revision.

\subsection{Stateful Adaptive Workflow}
\label{sec:stateful-workflow}

\begin{figure}[!htbp]
    \centering
    \includegraphics[width=\textwidth]{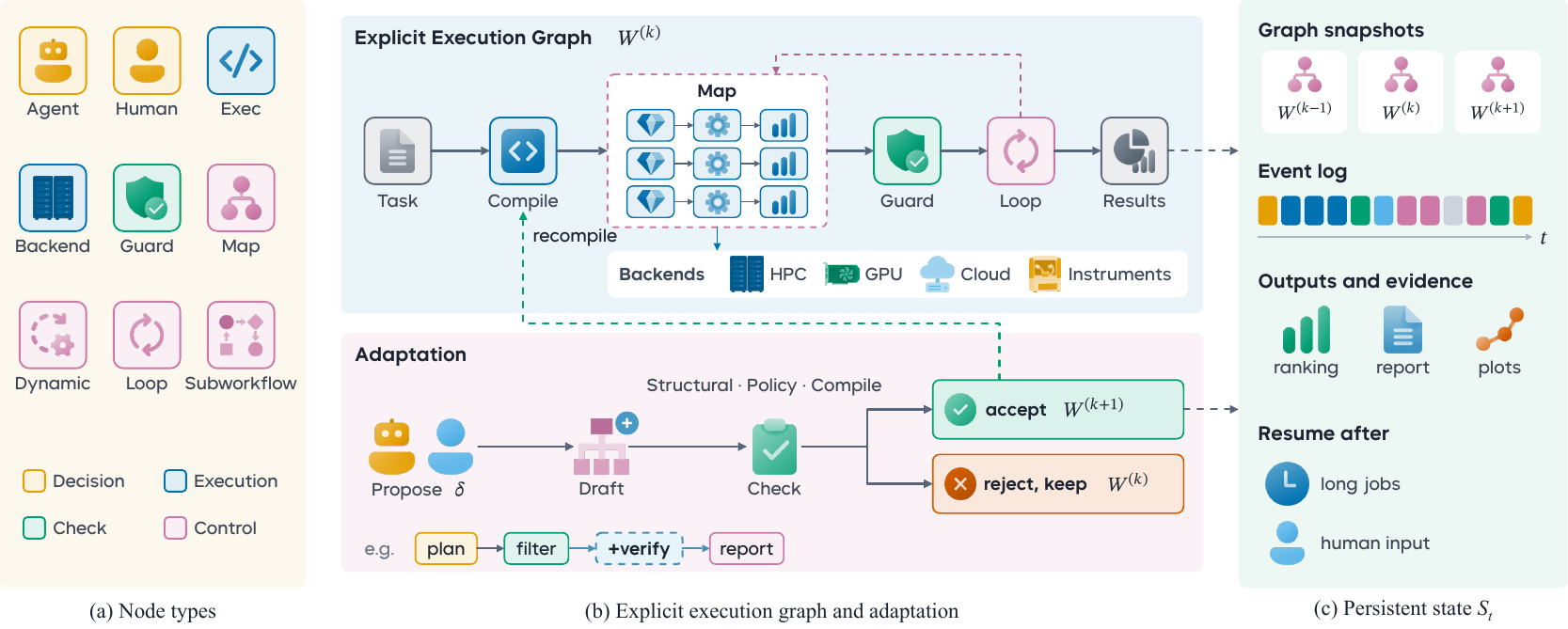}
    \caption{Stateful adaptive workflow. (a) Node types. (b) The execution graph (top) connects research tasks, workflow compilation, mapping, scientific checks, loops, and results. The adaptation path (bottom) evaluates graph updates proposed by an agent or researcher through structural, policy, and compilation checks. (c) Persistent state stores graph snapshots, event logs, and output evidence.}
    \label{fig:stateful-workflow}
\end{figure}

Language models can produce different outputs for identical inputs\cite{rabanser2026reliability}, and decisions in long tasks can drift as the available context changes. Written skills specify procedures but do not enforce their execution. MatOS therefore runs critical computations through an explicit execution graph that holds task steps, states, and checking rules (Figure~\ref{fig:stateful-workflow}), building on the dynamic workflows, error handling, and data provenance of materials workflow systems\cite{jain2015fireworks,huber2020aiida,ganose2025atomate2}. The runtime manages parameters, jobs, and evidence, while the agent revises plans and interprets anomalies.

\paragraph{Node types.}
A workflow decomposes a materials research task into nodes with explicit objectives, inputs, outputs, and completion conditions. The nine node types fall into four groups (Table~\ref{tab:node-mapping} and Figure~\ref{fig:stateful-workflow}a). Decision nodes (Agent and Human) assign open questions to a research role or a researcher. Execution nodes (Exec and Backend) run computations; Backend nodes submit jobs to external computing platforms and instruments through a common interface. Check nodes (Guard) test inputs and intermediate results against scientific or procedural conditions. Control nodes (Subworkflow, Map, Loop, and Dynamic) reuse processes, expand candidate batches, iterate, and propose changes to later steps. We denote the $k$-th revision of the execution graph by $W^{(k)}=(N^{(k)},A^{(k)})$, with nodes $N^{(k)}$ and control and dependency edges $A^{(k)}$; each node also records its asset references, input and output requirements, and transition conditions.

\begin{table}[htbp]
    \centering
    \small
    \caption{Node types and scientific responsibilities in the stateful adaptive workflow.}
    \label{tab:node-mapping}
    \begin{tabular}{@{}lp{0.68\textwidth}@{}}
        \toprule
        Node type & Scientific responsibility \\
        \midrule
        Agent & Assigns open decisions such as problem decomposition, method selection, and anomaly interpretation to a research role. \\
        Human & Handles stages requiring researcher judgment, additional information, or confirmation. \\
        Exec & Runs a specified computation, data-processing operation, or analysis step. \\
        Backend & Organizes a research job on an external computing platform or experimental instrument. \\
        Guard & Checks inputs and intermediate results against scientific or procedural conditions for continuation. \\
        Subworkflow & Invokes a reusable research process as a complete subtask. \\
        Map & Expands similar subtasks across candidate materials or research conditions. \\
        Loop & Organizes bounded iteration with repetition limits and stopping conditions. \\
        Dynamic & Proposes revisions to subsequent research steps and dependencies as a task progresses. \\
        \bottomrule
    \end{tabular}
\end{table}

\paragraph{Execution state and scheduling.}
The runtime repeats a cycle of checking prerequisites, running a node, recording its result, and advancing to its successors. The execution state $S_t$ at time $t$ holds the index $k_t$ of the current graph, the progress and results of each node, resource use, and process records. Whenever a node finishes, an external job returns, or a researcher provides feedback, the runtime updates $S_t$ and decides which steps can proceed.

The runtime supports two execution modes. In directed acyclic graph mode, the scheduler dispatches a node once its predecessors, inputs, and branch conditions are satisfied, then follows success, failure, or conditional branches according to the result. In state-machine mode, execution advances by research stage: the result of each node and explicit transition conditions determine the next node, and transitions back to an earlier stage implement feedback loops. In both modes, structured outputs are checked against the declared schema of their node, and each artifact is linked to the node and attempt that produced it, so the evidence remains available after later graph revisions. Before advancing, the runtime checks pause, cancellation, human confirmation, and budget conditions. Failures follow defined retry and failure routes, and tasks waiting for an external result or human input keep their progress until they can resume.

\paragraph{Long-running execution.}
Materials research spans widely varying timescales: a first-principles calculation may run for hours or days, and synthesis and characterization often have longer feedback cycles. Execution state therefore persists beyond the agent's active session: graph snapshots, node records, and an event log store $S_t$ independently of the model's context. For a long-running job, the agent hands execution to a background process and is notified when the run finishes, needs input, or stops responding, so it does not need to check the run repeatedly.

\paragraph{Checked graph updates.}
Predefined branches and loops cover anticipated changes. When a new result calls for changing later research steps, the agent or a researcher proposes a graph update $\delta$ based on the task objective, current state, and results, for example adding a validation node and making subsequent analysis depend on it. The proposal specifies added or modified nodes, dependency edges, and output requirements. It is admitted only if it passes three checks: a structural check of graph structure, asset references, restricted fields, and size bounds; a policy check of whether the operation is permitted under configuration $c$; and a compilation check that the revised definition yields a valid runtime graph. An admitted update is compiled into $W^{(k+1)}$, which becomes a new snapshot for subsequent scheduling; otherwise, the current graph continues to run. The runtime records each proposal, its checks, and the decision.

\subsection{Evidence-Driven Asset Evolution}
\label{sec:asset-evolution}

\begin{figure}[!htbp]
    \centering
    \includegraphics[width=\textwidth]{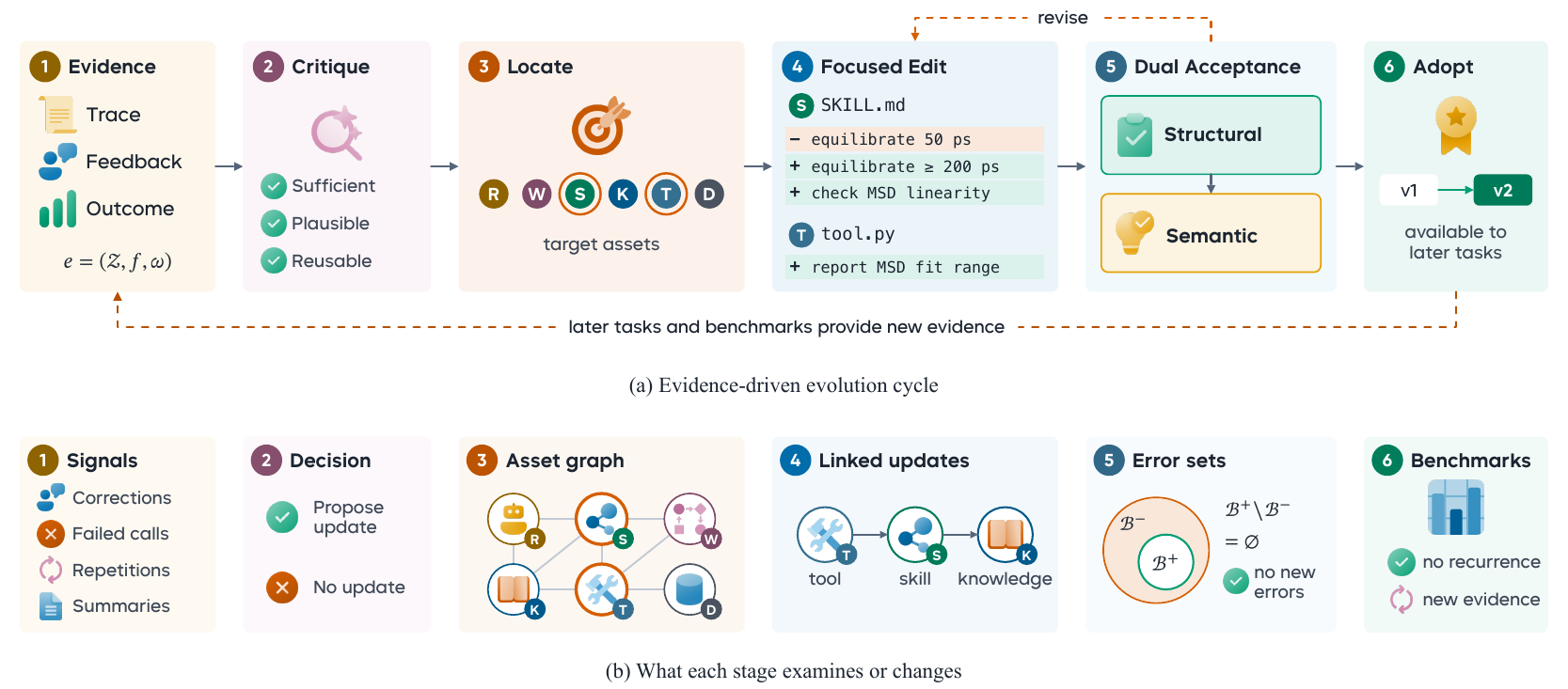}
    \caption{Evidence-driven asset evolution. (a) Evidence critique identifies assets for revision from execution traces and user feedback. Focused edits undergo structural checks and semantic review before adoption. Benchmarks and later tasks supply further feedback. (b) What each stage examines or changes, numbered as in (a).}
    \label{fig:asset-evolution}
\end{figure}

Task records and user feedback reveal reusable methods, failure causes, computing-environment requirements, and user preferences. As in memory- and skill-based approaches to agent self-improvement\cite{shinn2023reflexion,zhao2024expel,wang2024voyager}, MatOS records this experience outside the model and keeps the foundation-model parameters fixed. In MatOS, a single revision may span several linked assets and is adopted only after structural checks and semantic review. The process has six stages: evidence collection, critique, target location, focused editing, dual acceptance, and adoption (Figure~\ref{fig:asset-evolution}).

\paragraph{Evidence critique.}
User corrections, failed tool calls, repeated operations, task summaries, and workflow results provide candidate learning signals. MatOS packages each task and its feedback as evidence $e=(\mathcal Z,f,\omega)$, where $\mathcal Z$ is the execution trace, $f$ the user feedback, and $\omega$ the workflow outcomes and recovery records. A critic then asks whether the evidence is sufficient, whether the identified problem is plausible, whether the experience is reusable, and whether its applicability and risks are understood. Based on this critique, MatOS decides whether to propose an update and separates general method knowledge from individual preferences.

\paragraph{Target location and focused editing.}
Target location maps a signal that passes critique to the assets responsible for the relevant scientific function. MatOS first takes the assets named explicitly in the evidence, then adds related candidates ranked by their connections in the extended view $\mathcal G_c^{\mathrm{view}}$ (Section~\ref{sec:asset-graph}) and by textual relevance, keeping at most $m$ assets that are editable under configuration $c$. The appropriate target depends on the source of the problem: a method-selection error may call for revised knowledge or skills, a tool-behavior error for a repaired solver capability, and a missing research step for a workflow change. Because assets depend on one another, one change may require coordinated edits to related assets; after a tool interface changes, for example, skill steps and checking knowledge may need adjustment. An editor then produces a candidate update $\Delta$, which may change several existing assets or, when the evidence supports it, create a new one. A rejected candidate can be revised against the stated reasons a bounded number of times.

\paragraph{Dual acceptance.}
A candidate must be both structurally valid and semantically appropriate. Deterministic checks first test asset descriptions, interfaces, and composition relations. When a baseline from before the update is available, the same checks are run on the original and candidate states, and equivalent errors are matched across the two runs. Writing $\mathcal B^-$ and $\mathcal B^+$ for the errors found before and after the edit, the structural check passes when \mbox{$\mathcal B^+\setminus\mathcal B^-=\emptyset$}, that is, when the edit introduces no new error; errors that predate the edit do not block acceptance. A model-based semantic reviewer then examines the evidence, the critique, the candidate, and the check record, and asks whether the revision addresses the original evidence, preserves the method's applicability conditions, and avoids harming related capabilities. The candidate must also respect the edit permissions and data-use restrictions of configuration $c$, and it must keep linked assets consistent: a change to one asset must not contradict what related assets record about it. Critique, editing, and review are performed by separate components, and only candidates that pass all checks are adopted and become available to later tasks.

Subsequent tasks and benchmarks that use the edited assets show whether the original problem recurs, whether linked methods still work, and whether the experience transfers to related tasks. Their outcomes serve as evidence in the next round of critique.

\section{Experimental Evaluation}
\label{sec:benchmarks}

We evaluate MatOS at three levels: end-to-end research on ResearchClawBench, materials tool use on MatTools, and multistep scientific problem solving on SciAgentGym (Table~\ref{tab:benchmarks}). ResearchClawBench is the primary benchmark and reports execution cost alongside research quality.

\begin{table}[!htb]
\centering
\footnotesize
\setlength{\tabcolsep}{4pt}
\renewcommand{\arraystretch}{1.25}
\caption{Benchmarks used in this report.}
\label{tab:benchmarks}
\begin{tabular*}{\textwidth}{@{\extracolsep{\fill}}llll@{}}
\toprule
\textbf{Benchmark} & \textbf{Type} & \textbf{Tasks used} & \textbf{Metric} \\
\midrule
ResearchClawBench\cite{xu2026researchclawbench} & End-to-end research & 12 tasks & Rubric score, cost per task \\
MatTools\cite{liu2025mattools} & Materials software use & 49 problems, 138 subtasks & Subtask and problem success \\
SciAgentGym\cite{shen2026sciagentgym} & Multistep tool use & 42 cases & Case success rate \\
\bottomrule
\end{tabular*}
\end{table}

\subsection{Benchmarks}
ResearchClawBench evaluates end-to-end autonomous research using tasks derived from published papers\cite{xu2026researchclawbench}. A system receives relevant literature and raw data while the target paper remains hidden. It must independently understand the question, execute experiments and computations, and write a research report. Domain experts select and construct the tasks, and the final report is scored on evaluation items, weighted by importance, that cover the quantitative results, figure evidence, and scientific analysis of the target research. We use the chemistry, energy, and materials tasks, which include molecular modeling, battery parameters, energy-system analysis, atomistic models, and inverse materials design. Each task spans data preparation, model execution, and scientific interpretation.

MatTools tests understanding and use of materials science software\cite{liu2025mattools}. We use its real-world tool-use component, comprising 49 problems and 138 property-computation subtasks. The problems involve pymatgen\cite{ong2013pymatgen} and its defect-analysis extension\cite{shen2024pmgdefects} and cover charge-density analysis, defect construction, interstitial-site identification, and related property calculations. From the question and the software documentation, a system must generate an executable function that returns the prescribed materials properties and data structures and passes unit tests, which check executability and the correctness of the materials results separately. Correct results require translating materials concepts into software: the defect type determines the relevant classes and methods, and crystal coordinates and symmetry equivalence determine the structure of the returned objects.

SciAgentGym provides an interactive scientific environment with 1,780 domain tools\cite{shen2026sciagentgym}. Its tasks range from basic operations to multistep tool composition, with tool outputs and execution feedback informing later reasoning. Some problems require interpretation of molecular structures, diffraction figures, or spectral images. We use a fixed set of 42 public chemistry and materials cases: 22 chemistry cases and 20 materials-related cases involving molecular property calculations, diffraction, and spectral analysis. Each case requires a solution path from observations or structures to a scientific answer, with software outputs interpreted under the question's physical constraints.

\subsection{Experimental Setup}
The MatOS asset library was frozen throughout the evaluation, so no benchmark task contributed to asset revisions. Appendix~\ref{app:model-versions} lists the model identifiers, settings, and dates of all runs. On ResearchClawBench, we evaluate MatOS with DeepSeek-V4-Flash\cite{deepseekai2026v4} and GPT-5.4\cite{openai2026gpt54}. We score submissions with the benchmark's official rubric, aggregate the evaluation items of each task, and report domain means and the overall mean over 12 tasks. Baselines are MIRA\cite{mira2026general}, ResearchClaw\cite{researchclaw2026software}, EvoScientist\cite{lyu2026evoscientist}, Codex CLI\cite{openai2026codexcli}, OpenClaw\cite{openclaw2026software}, ARIS Codex\cite{openai2026codexcli,yang2026aris}, and Nanobot\cite{hkuds2026nanobot}. All baselines use GPT-5.4.

MatTools and SciAgentGym compare MatOS, Codex CLI\cite{openai2026codexcli}, and Claude Code\cite{anthropic2025claudecode}, using DeepSeek-V4-Pro\cite{deepseekai2026v4} as the base model for all systems. Holding the model fixed helps separate the effect of the agent system from that of the model. All systems receive the same task inputs and produce artifacts in independent working directories. MatOS runs with its compute role, which selects among the available domain tools and skills. On MatTools, the same deterministic adapter converts the output of each system into the format expected by the unit tests before the tests are run. We report success over 138 property subtasks and 49 complete problems. A problem is complete when all its subtasks pass. On SciAgentGym, each system's final answers are graded case by case against a common rubric, and a case counts as a success only if the answer is wholly correct.

\subsection{Results and Discussion}

\begin{figure}[!htb]
    \centering
    \includegraphics[width=\linewidth]{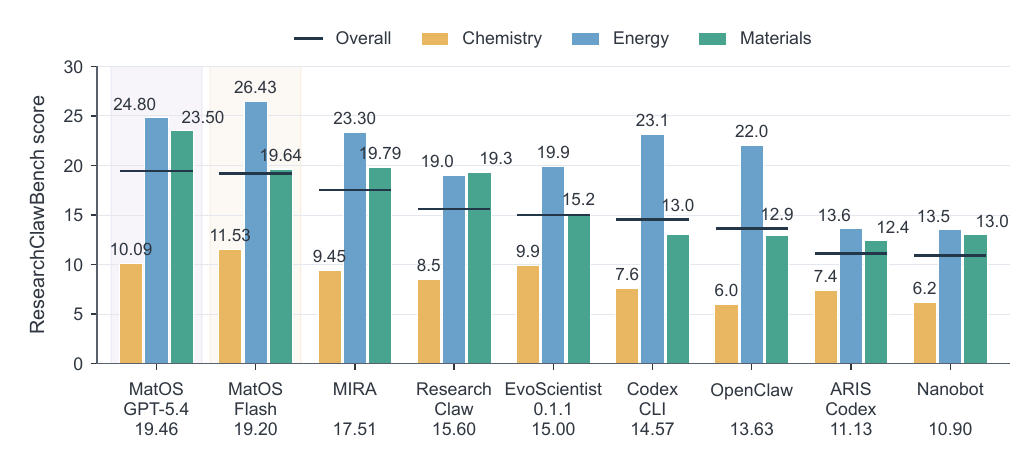}
    \caption{ResearchClawBench domain and overall scores. Bars show the chemistry, energy, and materials scores; horizontal markers show the overall score, also listed below each system name.}
    \label{fig:rcbench-scores}
\end{figure}

\paragraph{End-to-end research quality and cost.}
Figure~\ref{fig:rcbench-scores} compares the domain scores. MatOS scores 19.46 overall with GPT-5.4 and 19.20 with DeepSeek-V4-Flash. With the same base model as the baselines, the GPT-5.4 configuration exceeds the highest-scoring baseline, MIRA, by 11.15\%; the lower-cost DeepSeek-V4-Flash configuration exceeds it by 9.63\%. In chemistry, the two configurations score 10.09 and 11.53, whereas no baseline exceeds 9.90 (EvoScientist). In energy, they score 24.80 and 26.43, compared with 23.30 for MIRA. In materials, the GPT-5.4 configuration scores 23.50, 18.7\% above MIRA's 19.79.

\begin{figure}[!htb]
    \centering
    \includegraphics[width=\linewidth]{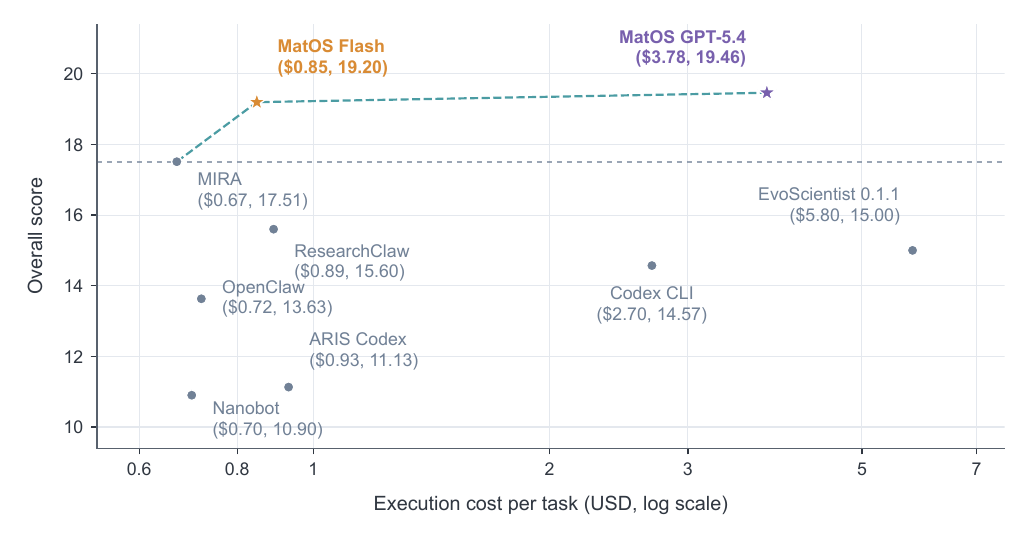}
    \caption{ResearchClawBench mean execution cost versus overall score. Cost is shown on a logarithmic axis. The teal dashed line joins the observed non-dominated points, and the gray dashed line marks the score of MIRA, the highest-scoring baseline.}
    \label{fig:rcbench-cost}
\end{figure}

Figure~\ref{fig:rcbench-cost} plots cost against score. Among the baselines, only MIRA lies on the observed non-dominated front. The two MatOS configurations, at $(\$0.85,19.20)$ and $(\$3.78,19.46)$, extend this front: the DeepSeek-V4-Flash configuration scores higher than MIRA at a cost close to MIRA's (\$0.85 vs.\ \$0.67), and the GPT-5.4 configuration achieves the highest score at a higher cost.

\FloatBarrier

A representative artifact shows what a high-scoring submission contains. On the materials task \texttt{Material\_002}, which targets an adsorption scaling relation, the GPT-5.4 configuration scores 34.14 by building the full computational pipeline itself. It uses ASE\cite{larsen2017ase} and MACE\cite{batatia2022mace} to construct fcc(111) surfaces of Ni, Cu, Rh, Pd, Ir, and Pt, relaxes the clean and O- or OH-adsorbed surfaces, computes adsorption energies with consistent gas-phase references, and fits a linear relation across the six metals. The submitted slope is 0.758 with $R^2=0.960$, and the rubric credits the submission with reproducing the main scaling trend of the target study.

\begin{figure}[!htb]
    \centering
    \includegraphics[width=\linewidth]{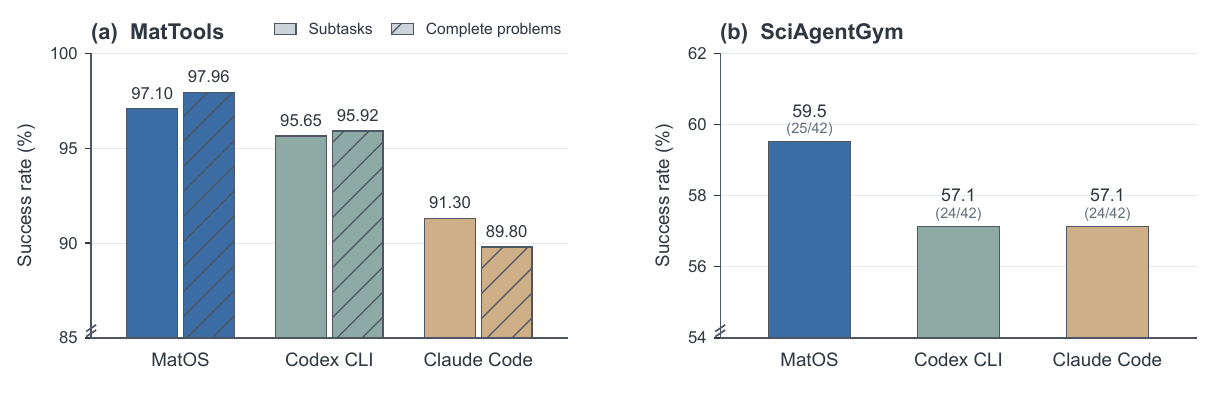}
    \caption{Materials tool and scientific-task results. (a) MatTools subtask success (solid fill) and complete-problem success (hatched). (b) SciAgentGym case success, with solved-case counts in parentheses.}
    \label{fig:tool-benchmark-comparison}
\end{figure}

\paragraph{Materials tool use and complete submissions.}
In Figure~\ref{fig:tool-benchmark-comparison}(a), MatOS passes 134 of 138 subtasks (97.10\%) and 48 of 49 complete problems (97.96\%), compared with 132 and 47 (95.65\% and 95.92\%) for Codex CLI and 126 and 44 (91.30\% and 89.80\%) for Claude Code. Relative to Codex CLI, MatOS improves complete-problem success by 2.04 percentage points, a relative gain of 2.13\%. Every generated function from all three systems imports and runs, so the differences come from property computation and result structure.

The three systems differ on only five problems, each of which turns on how a materials result is represented. Claude Code fails four of them. In the Fe$_3$O$_4$ charge-density task \texttt{test\_chgcar\_insertion}, it calls the same insertion analyzer as MatOS but flattens the symmetry-equivalent site groups, which loses the mapping between mean charge and candidate sites; in vacancy, interstitial, and multi-defect tasks, it returns one to three incorrect properties. Codex CLI fails one problem, \texttt{test\_wswq\_slope}, by assigning two series of configuration-coordinate distortions to the opposite slope properties. MatOS passes all five.

\FloatBarrier

\paragraph{Scientific tool composition and numerical answers.}
Figure~\ref{fig:tool-benchmark-comparison}(b) shows a case success rate of 59.5\% for MatOS (25 of 42), compared with 57.1\% (24 of 42) for both Codex CLI and Claude Code, a relative gain of 4.17\%. Unlike MatTools, these tasks involve several tools, and a system must carry physical quantities, units, and structural conditions from one tool call to the next.

For a molecular rigidity-index task, MatOS calls the benchmark's geometry-optimization and three-dimensional-descriptor tools on norbornane, norbornene, and norbornadiene. The tools generate conformers with the ETKDGv3 method in RDKit\cite{wang2020etkdgv3,rdkit} and optimize them with the MMFF94 force field; MatOS then applies the specified formula to the radius of gyration, the ratio of principal moments of inertia, and the eccentricity. The submission keeps the individual descriptors and rigidity indices and obtains a mean of 82.28, judged wholly correct. Codex CLI and Claude Code instead compute the descriptors with their own RDKit scripts, obtain different descriptor values, and both report 81.12, which receives partial credit.

\paragraph{Observations across benchmarks.}
Across the three benchmarks, the decisive steps keep physical objects and their software representations consistent from one computation to the next: symmetry-equivalent site groups and the sign of a distortion in MatTools, the descriptor route in SciAgentGym, and consistent gas-phase references in ResearchClawBench. On MatTools and SciAgentGym, where all three systems solve most problems, the margins come from a few problems of this kind. On open-ended research tasks, where such steps accumulate over a whole study, the gain over the previously best-performing agent is larger, at 11.15\%. MatOS records these conventions in its knowledge and skill assets and routes methods through its roles, so that an agent can reuse a checked procedure instead of reconstructing it for each task.

\section{Research Cases}
\label{sec:scientific-cases}

MatOS was applied to eight research cases that span several classes of materials. Section~\ref{sec:cases-potentials} begins with interatomic potentials across inorganic materials, which underlie many of the simulations in the later cases: MatOS evaluates 11 pretrained potentials on metallic, oxide, and other compound configurations and helps develop a training-data selection method. The following sections turn to specific material systems: inorganic solid electrolytes (Section~\ref{sec:cases-solid-electrolytes}), liquid electrolytes (Section~\ref{sec:case-liquid-electrolyte}), and alloys (Section~\ref{sec:cases-alloys}), with tasks ranging from large-scale screening and molecular simulation to the development of a prediction model. Appendix~\ref{app:cfdbench-cavity} reports the eighth case, a reproduction of the CFDBench lid-driven cavity benchmark in continuum fluid dynamics.

\subsection{Interatomic Potentials Across Inorganic Materials}
\label{sec:cases-potentials}

\subsubsection{Error Analysis of Pretrained Potentials}
\label{sec:case-mlip-benchmark}

\begin{figure}[!htb]
    \centering
    \includegraphics[width=\linewidth]{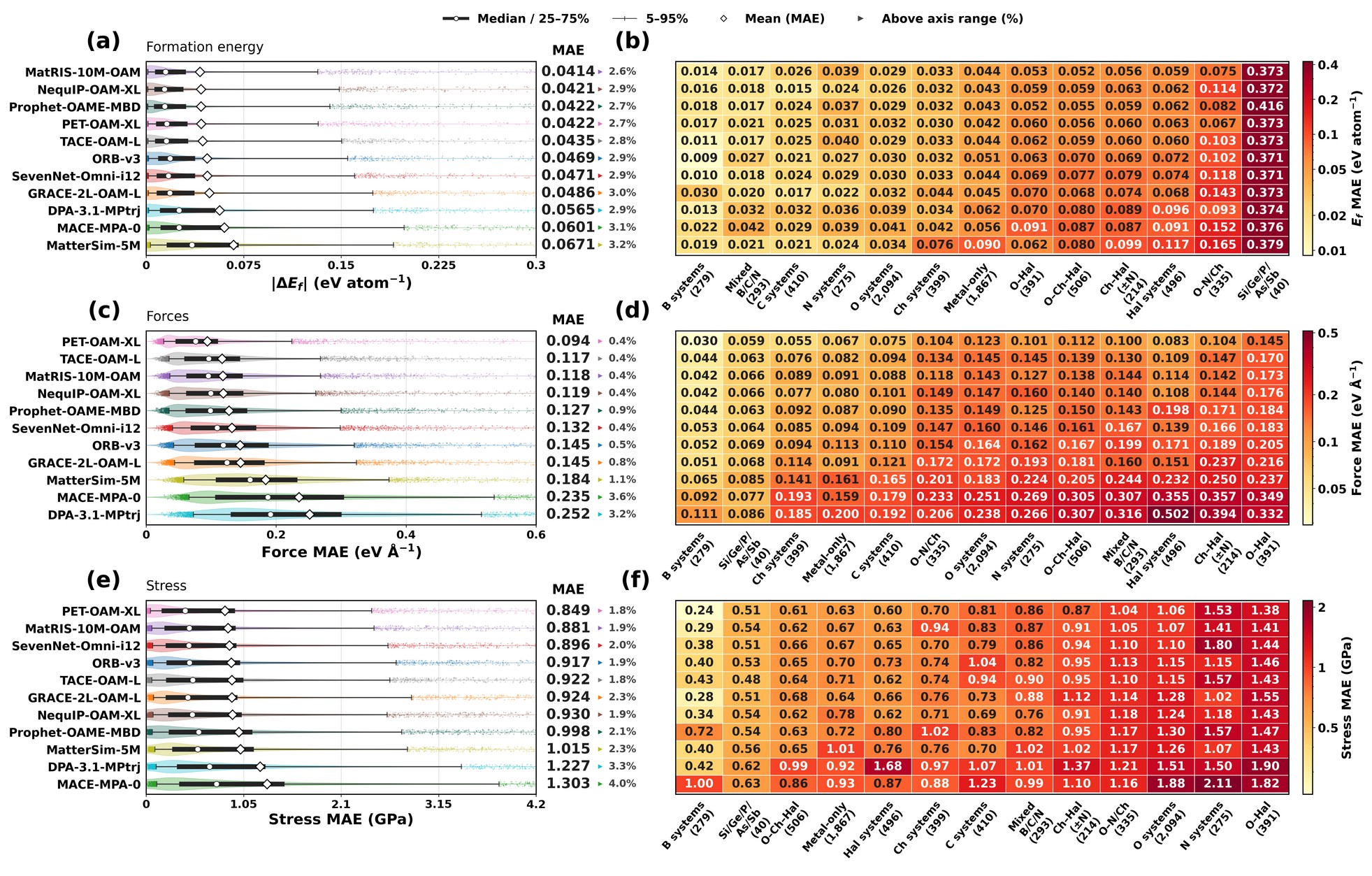}
    \small
    \caption{Energy, force, and stress errors of 11 pretrained potentials on 7,599 multicomponent configurations. (a,c,e) Structure-level distributions of formation-energy error, Cartesian force MAE, and six-component stress MAE. Symbols and intervals denote the mean, median, interquartile range, and 5th to 95th percentiles. (b,d,f) Mean errors within disjoint chemical classes. Rows and columns are ordered by model and class mean errors for each property. Formation energies use aligned model and DFT elemental references. Ch denotes S, Se, or Te, and Hal denotes F, Cl, Br, or I. Adapted from Du \emph{et al.}\cite{du2026originsuniversalmachinelearning}.}
    \label{fig:case-mlip-benchmark}
\end{figure}

MatOS evaluates 11 pretrained interatomic potentials on a set of 7,599 multicomponent configurations\cite{du2026originsuniversalmachinelearning}, comprising 1,867 metallic or intermetallic, 3,326 oxygen-containing, and 2,406 oxygen-free compound configurations (Figure~\ref{fig:case-mlip-benchmark}). All models use the same structures and DFT labels. Formation energies are compared after aligning elemental references, while forces and stresses are compared directly with DFT. MatRIS-10M-OAM\cite{zhou2026matris} gives the lowest mean formation-energy error, 0.0414~eV\,atom$^{-1}$. Structure-averaged force MAE ranges from 0.0941~eV\,\AA$^{-1}$ for PET-OAM-XL\cite{mazitov2025petmad,upet} to 0.2516~eV\,\AA$^{-1}$ for DPA-3.1-MPtrj\cite{zhang2026dpa3}. Mean stress MAE ranges from 0.849 to 1.303~GPa, with PET-OAM-XL again giving the lowest value. High force errors recur across models in oxygen--halogen, chalcogen--halogen, and halogen systems.

\begin{figure}[!htb]
    \centering
    \includegraphics[width=\linewidth]{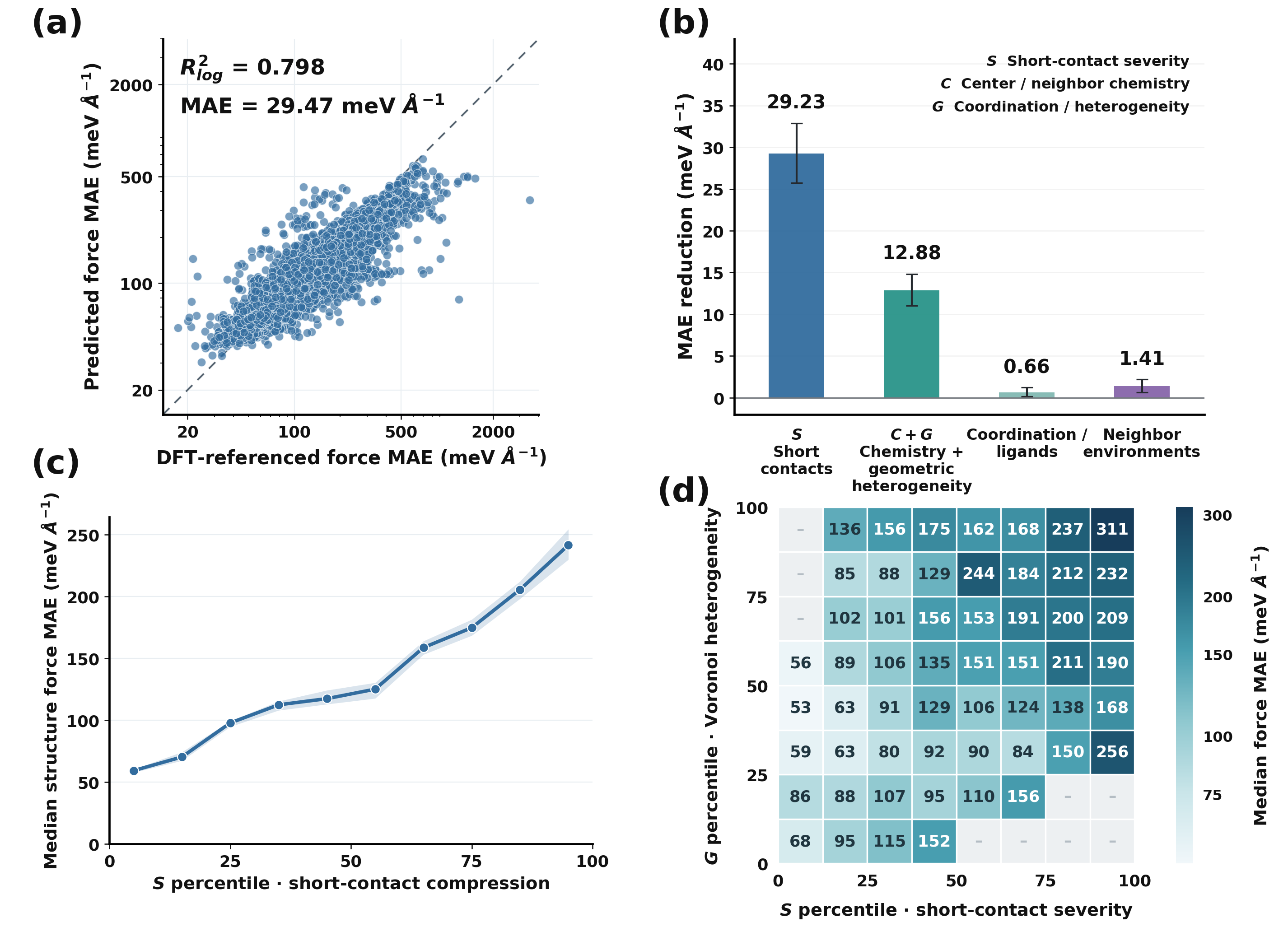}
    \small
    \caption{Local chemical and geometric contributions to force-error prediction. (a) Out-of-fold predictions of the mean force error across 11 potentials, averaged to the structure level for 7,599 configurations. (b) Reductions in prediction MAE from a constant baseline to a short-contact model, then to the full descriptor model. Smaller bars show additional coordination and neighbor-environment descriptors. Error bars are 95\% intervals from 1,000 paired group-bootstrap resamples. (c) Median force error in ten percentile bins of short-contact severity, with 95\% bootstrap intervals. (d) Median force error versus short-contact severity and Voronoi heterogeneity for 7,583 valid structures. Grey cells contain fewer than 20 structures. Both coordinates are percentile ranks. $S$ denotes short-contact severity, $C$ central and neighboring chemistry, and $G$ coordination and geometric heterogeneity. Adapted from Du \emph{et al.}\cite{du2026originsuniversalmachinelearning}.}
    \label{fig:case-mlip-local}
\end{figure}

An atom-based gradient-boosting regressor relates local descriptors to the mean Cartesian force error across the 11 potentials (Figure~\ref{fig:case-mlip-local}). Five-fold grouped cross-validation keeps related structures together, and atomic predictions are averaged within each structure. Four short-contact descriptors give a prediction MAE of 42.35~meV\,\AA$^{-1}$. Adding chemical and geometric descriptors lowers this to 29.47~meV\,\AA$^{-1}$, a 30.4\% reduction, with $R^2_{\log}=0.798$. The absolute reduction is 12.88~meV\,\AA$^{-1}$, with a 95\% grouped-bootstrap interval of 11.04 to 14.81~meV\,\AA$^{-1}$. Across ten bins of increasing short-contact severity, the median force error rises from approximately 60 to 240~meV\,\AA$^{-1}$. Large errors concentrate where short contacts coincide with uneven neighbor distances, measured by Voronoi-face-area-weighted heterogeneity. A separate analysis matches element pairs and absolute distance deviations from their training-reference medians. Contacts shorter than the reference median have mean force errors 1.81 to 1.95 times those of equally displaced longer contacts. Local chemistry, short contacts, and coordination-shell geometry can therefore guide force-error estimation and selection of additional reference calculations.

\FloatBarrier
\subsubsection{Training-Data Selection}
\label{sec:case-mptrj-selection}

The task is to improve interatomic-potential accuracy at a fixed training-data budget. Using MatOS for strategy development and implementation, training coordination, and comparative error analysis, we developed RarityGuard-MP, a training-data selection method that uses geometric information from relaxation trajectories to prioritize rare local environments. Selection does not use candidate energy, force, or stress labels.

\begin{figure}[!htb]
    \centering
    \includegraphics[width=\linewidth]{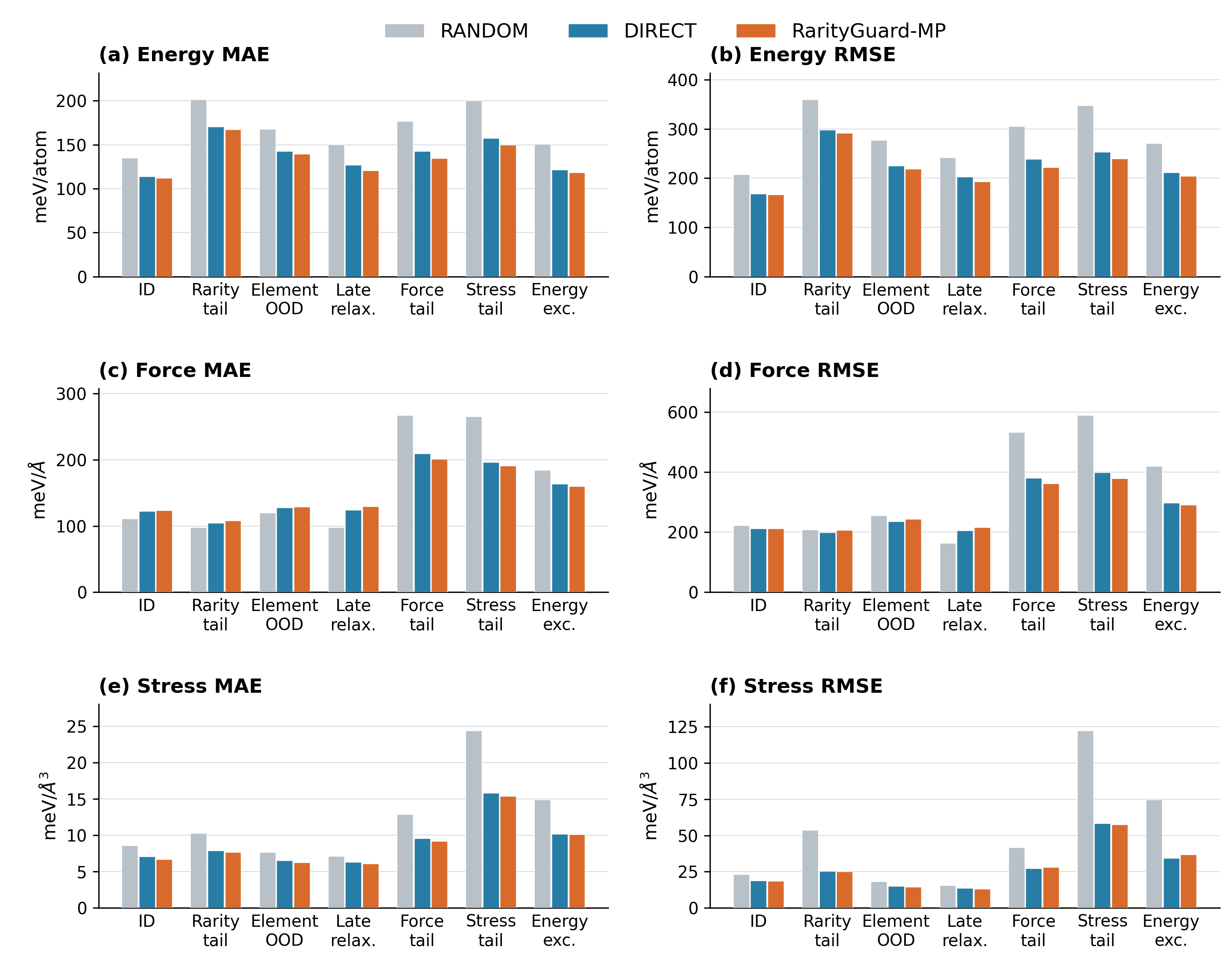}
    \small
    \caption{Allegro prediction errors for a single training seed (20260813) on seven MPtrj evaluation sets, with 100,000 training structures per selection method. Models were trained for 300 epochs with a global batch size of 512. (a,b) Energy MAE and RMSE. (c,d) Force-component MAE and RMSE. (e,f) Stress-component MAE and RMSE.}
    \label{fig:case-mptrj-selection}
\end{figure}

We compared RarityGuard-MP with DIRECT\cite{qi2024direct} and RANDOM on an MPtrj\cite{deng2023chgnet} pool of 1,263,071 structures, selecting 100,000 structures per method. Allegro\cite{musaelian2023allegro} models were trained from scratch under common training settings, with checkpoints selected on validation data. Seven evaluation sets cover in-distribution structures, rare chemistry, unseen element combinations, late relaxation, large forces, large stresses, and relaxation energy excursions. Energy, force, and stress MAE and RMSE provide 42 metrics in total. Aggregate performance is measured by the geometric mean of paired error ratios with equal metric weights. The evaluation sets were also used during selector development.

\begingroup
\clubpenalty=10000
\widowpenalty=10000
Across three Allegro training seeds, RarityGuard-MP has lower aggregate error than both baselines in every seed, with overall reductions of 3.10\% relative to DIRECT and 14.46\% relative to RANDOM. Relative to DIRECT, the aggregate energy, force, and stress error ratios are 0.9672, 0.9958, and 0.9448, respectively. Most of the improvement comes from energy and stress prediction; the aggregate force-error reduction is small.
\par\endgroup

Figure~\ref{fig:case-mptrj-selection} shows the Allegro comparison for a single training seed. For the three-seed comparison, the DIRECT and RarityGuard-MP selections are reused, while a new fixed RANDOM selection is sampled uniformly across lineages. All selections are held fixed across training seeds, enabling comparisons of prediction accuracy and training variability at a matched structure budget.

\FloatBarrier
\subsection{Inorganic Solid Electrolytes}
\label{sec:cases-solid-electrolytes}

\subsubsection{Solid-Electrolyte Screening}
\label{sec:case-sse-screening}

\begin{figure}[!htb]
    \centering
    \includegraphics[width=\linewidth]{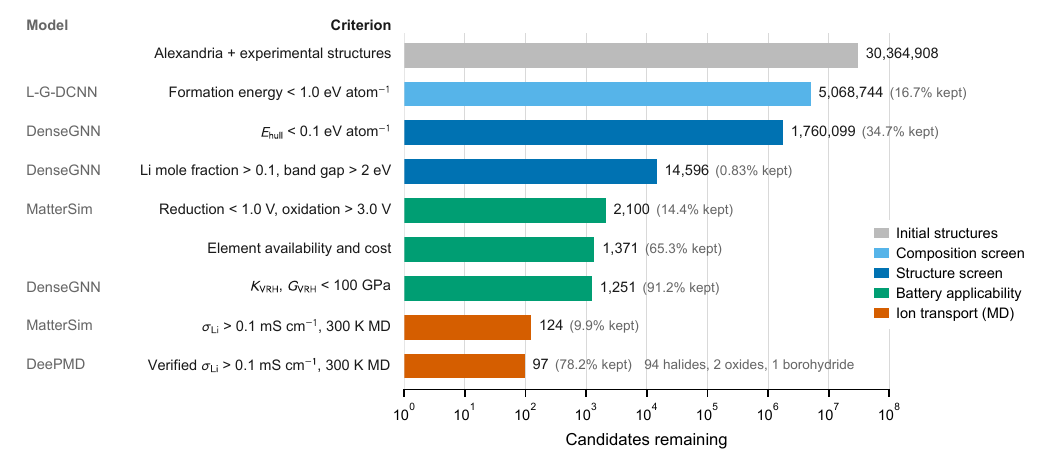}
    \small
    \caption{Staged screening of solid electrolytes. Bars show the remaining candidate count on a logarithmic scale, with colors indicating screening stages. Models are listed at left, and percentages give the fraction retained from the previous stage. The screen narrows 30,364,908 initial structures to 97 candidates.}
    \label{fig:case-sse-screening}
\end{figure}

\begin{figure}[!htb]
    \centering
    \includegraphics[width=\linewidth]{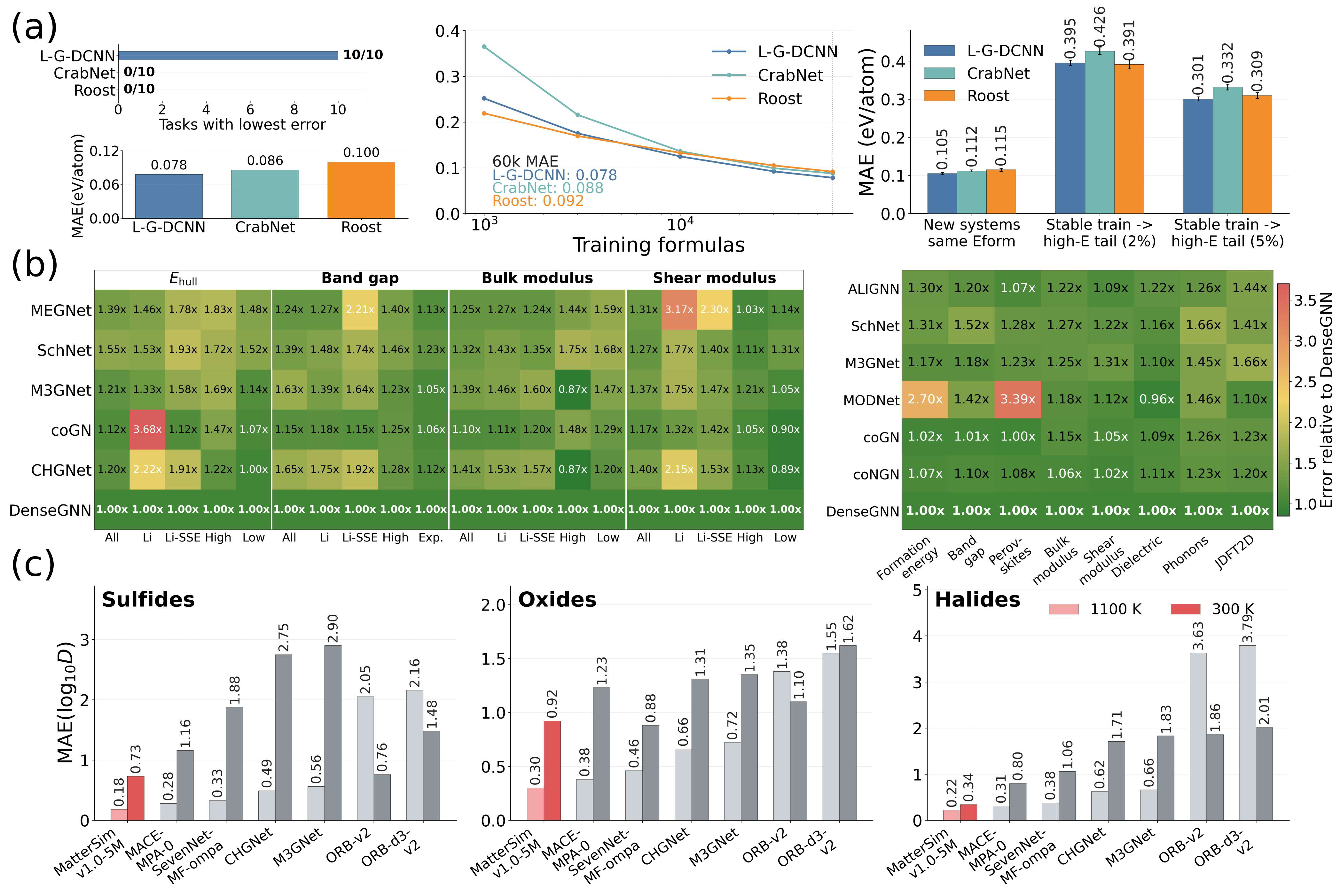}
    \small
    \caption{Model benchmarks for hierarchical solid-electrolyte screening. (a) Composition-based L-G-DCNN evaluation: counts of lowest-MAE Matbench tasks, formation-energy prediction and learning curves, and three composition-only extrapolation tests. Extrapolation error bars show standard deviations across five folds. (b) Structure-based property prediction and Matbench comparisons, with MAEs normalized to DenseGNN. (c) Mean absolute errors in $\log_{10} D$ for Li$^+$ diffusion in sulfides, oxides, and halides at 1100 and 300~K, using system-specific DeePMD simulations as references. Adapted from Du \emph{et al.}\cite{du2026ssehier}.}
    \label{fig:case-sse-benchmarks}
\end{figure}

MatOS screens solid electrolytes for ionic transport, electronic insulation, electrochemical stability, and mechanical compatibility\cite{famprikis2019sse} using the hierarchical framework of Du \emph{et al.}\cite{du2026ssehier}. Model benchmarks guide the choice at each level (Figure~\ref{fig:case-sse-benchmarks}). L-G-DCNN achieves a formation-energy MAE of 0.078~eV\,atom$^{-1}$, while DenseGNN provides the reference accuracy for comparisons of stability, band gap, and elastic-property predictions. Against system-specific DeePMD diffusion references, MatterSim has the lowest mean absolute logarithmic error across sulfides, oxides, and halides at 1100~K and across sulfides and halides at 300~K. For oxides at 300~K, its error is 0.92, compared with 0.88 for SevenNet-MF-ompa. Starting from 30,364,908 candidate structures derived from Alexandria\cite{schmidt2024alexandria} and experimentally determined inorganic crystal structures\cite{du2026ssehier}, the workflow applies composition models, graph neural networks, and molecular dynamics in sequence, reserving transport simulations for candidates that pass the preceding property filters (Figure~\ref{fig:case-sse-screening}).

L-G-DCNN, which combines long short-term memory and gated recurrent unit layers with a deep convolutional neural network\cite{du2024lgdcnn}, first screens compositions without structural input and retains 5,068,744 candidates. DenseGNN\cite{du2024densegnn} then predicts stability, band gap, and elastic properties, leaving 14,596 structures after lithium-content criteria are applied. Electrochemical-window, elemental-availability, and mechanical constraints reduce the set to 1,251. MatterSim\cite{yang2024mattersim} molecular dynamics at 300~K identifies 124 structures with potentially fast ion conduction. DeePMD\cite{zeng2023deepmd} trajectory review and quality checks leave 97 candidates: 94 halides, two oxides, and one borohydride, with calculated conductivities of 0.109 to 59.0~mS\,cm$^{-1}$. Of the 94 halides, 76 fall within the high-conductivity structural regions identified by Li \emph{et al.} from experimental data in a cation-polarization descriptor space\cite{li2024halide}. Analysis of the trajectories relates conductivity to migration-path connectivity: Li$_6$CaCl$_8$ forms channels connected across supercells, whereas Li$_3$BO$_3$ shows only a few local hops in the available trajectory.

\FloatBarrier
\subsubsection{Halide Electrolyte Variants}
\label{sec:case-halide-adaptation}

\begin{figure}[!htb]
    \centering
    \includegraphics[width=\linewidth]{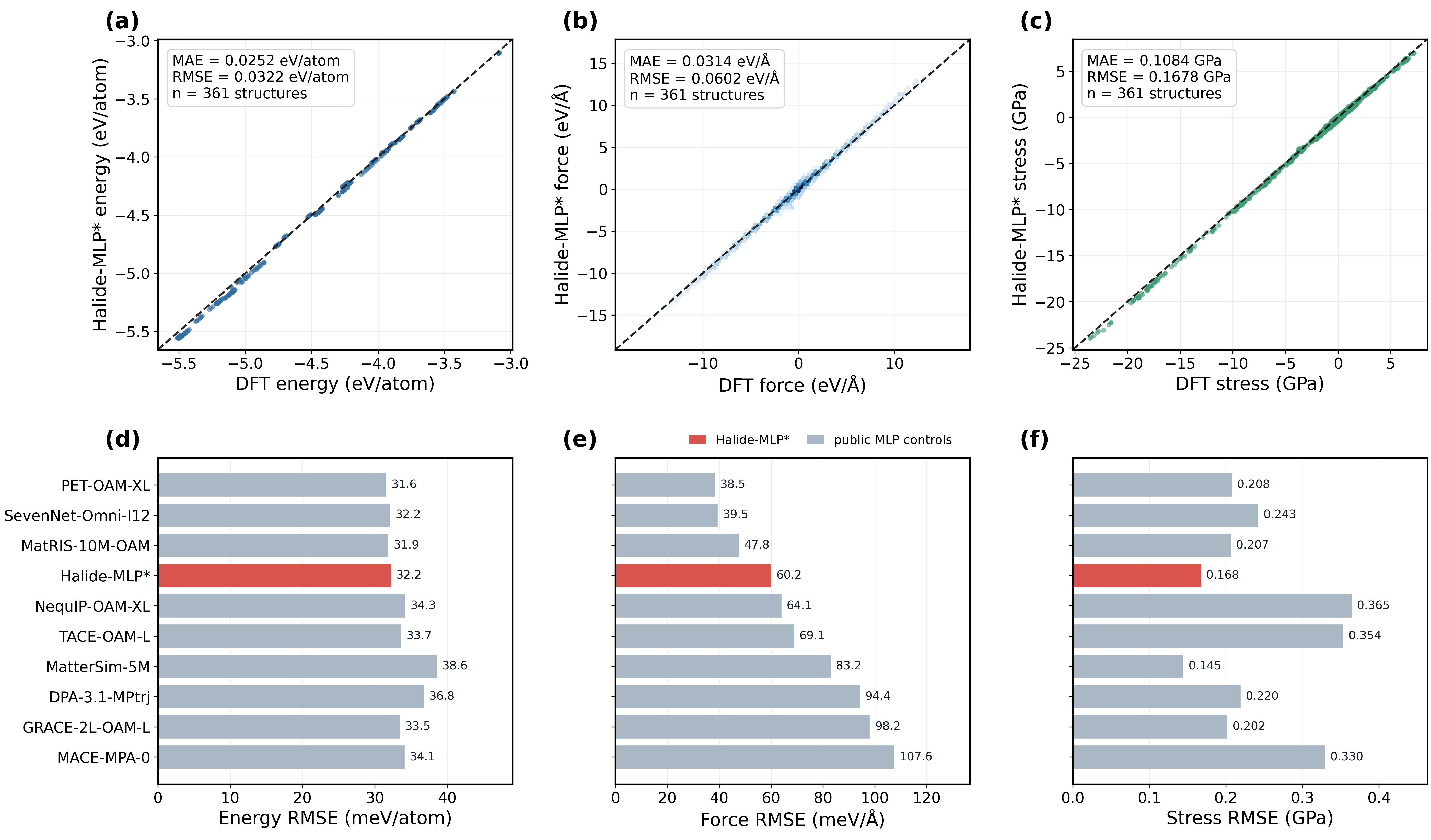}
    \small
    \caption{ORB-v3 predictions against DFT references for 361 halide configurations. (a--c) Energy, force, and stress parity plots. (d--f) Corresponding RMSEs compared with nine other public potentials. The source figure labels ORB-v3 as Halide-MLP*.}
    \label{fig:case-halide-validation}
\end{figure}

Starting from 23 low-conductivity halide structures, MatOS generates 2,211 variants through lithium defects, aliovalent substitutions, antisites, cation--anion disorder, and extended defects. SOAP novelty and disagreement among MACE models guide configuration selection for DFT labeling\cite{zhang2020dpgen}, producing a reference dataset of 10,019 structures. ORB-v3\cite{rhodes2025orbv3} supplies energies and forces for property screening. On a separate test set of 361 strained, sheared, and thermally perturbed configurations, its energy, force, and stress RMSEs are 32.2~meV\,atom$^{-1}$, 60.2~meV\,\AA$^{-1}$, and 0.168~GPa, respectively (Figure~\ref{fig:case-halide-validation}). For nine other public potentials evaluated on the same configurations, energy RMSEs range from 31.6 to 38.6~meV\,atom$^{-1}$. One has a lower stress RMSE than ORB-v3, and three have lower force RMSEs, ranging from 38.5 to 47.8~meV\,\AA$^{-1}$.

Successive property filters reduce the 2,211 variants to 1,862 with DFT $E_{\mathrm{hull}}\le0.10$~eV\,atom$^{-1}$, then 1,614 with reduction potentials below 1.0~V and oxidation potentials above 3.0~V. Requiring positive bulk and shear moduli below 100~GPa leaves 1,451 variants. Bond-valence site energy (BVSE)\cite{chen2019softbv} and nudged elastic band (NEB)\cite{henkelman2000neb} migration barriers of at most 0.50~eV reduce the set to 432, and PBE band gaps above 2.0~eV leave 394. Of these, 100 have completed 1~ns molecular dynamics simulations at 300~K. Eighty-nine reach conductivities of at least 0.1~mS\,cm$^{-1}$, exceed their parent structures, and pass DFT checks of energies, forces, and stresses.

\FloatBarrier
\subsection{Liquid Electrolytes}
\label{sec:case-liquid-electrolyte}

\begin{figure}[!htb]
    \centering
    \includegraphics[width=\linewidth]{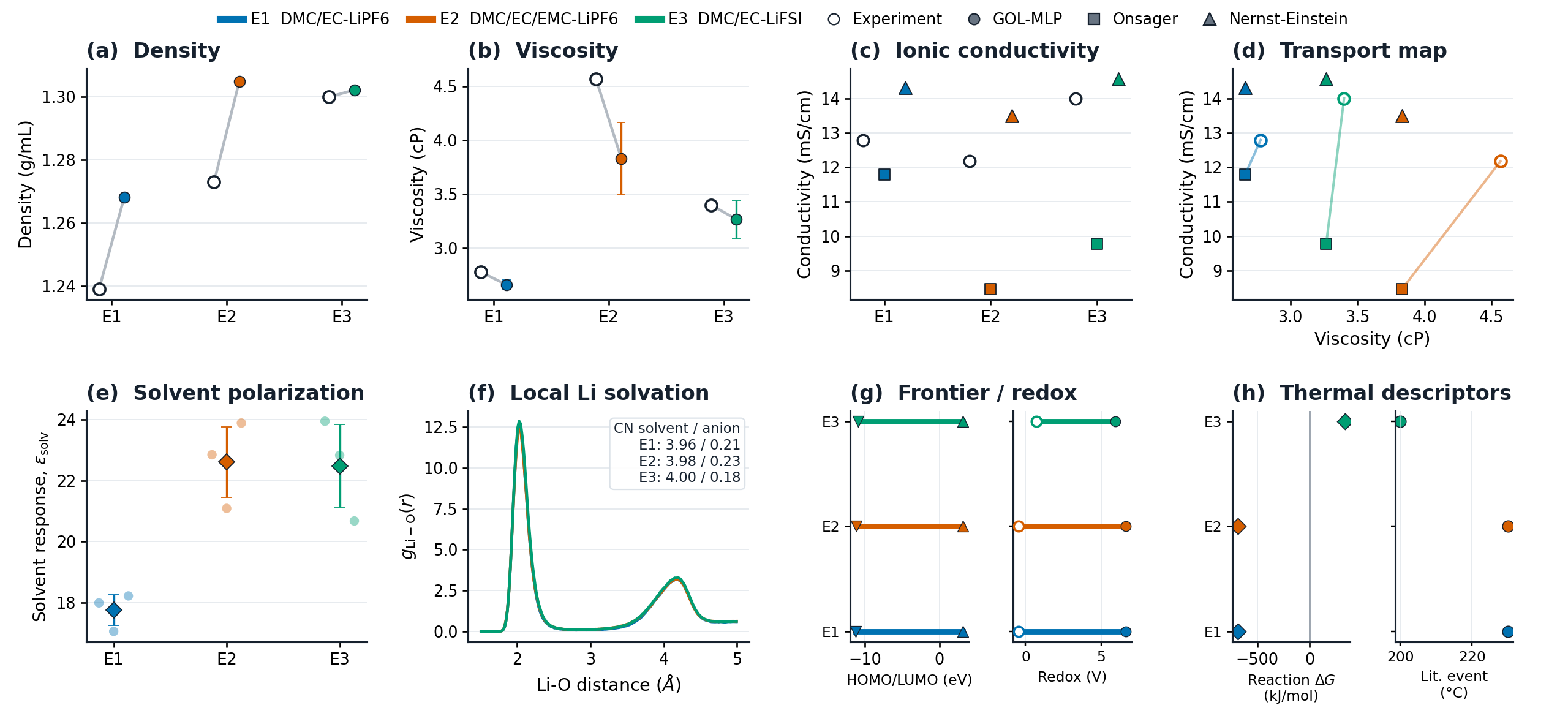}
    \small
    \caption{Calculated properties of liquid electrolytes E1, E2, and E3. (a--d) Density, viscosity, and conductivity from molecular dynamics. (e) Finite-field response of the neutral solvent. (f) Local Li$^+$ solvation from molecular dynamics. (g,h) Molecular quantum-chemical descriptors and thermal properties. Reaction free energies are calculated, while thermal-event temperatures are literature values.}
    \label{fig:case-liquid-electrolyte}
\end{figure}

MatOS compares solvation and transport in three liquid electrolytes\cite{xu2014electrolytes} at 298~K (Figure~\ref{fig:case-liquid-electrolyte}). E1 contains 0.9~M LiPF$_6$ in dimethyl carbonate (DMC) and ethylene carbonate (EC) at 60/40. E2 contains 1.1~M LiPF$_6$ in DMC/\allowbreak EC/\allowbreak ethyl methyl carbonate (EMC) at 45/50/5, and E3 contains 1.12~M lithium bis(fluorosulfonyl)imide (LiFSI) in DMC/EC at 51/49. Packmol\cite{martinez2009packmol} constructs periodic models of 10,046, 10,205, and 10,114 atoms, respectively. OpenMM\cite{eastman2024openmm} performs NPT equilibration and transport simulations using GOL-MLP, a polarizable graph-neural-network potential for organic liquids. ORCA\cite{neese2020orca} provides molecular quantum-chemical descriptors. Calculated densities of approximately 1.27 to 1.31~g\,mL$^{-1}$ are within about 3\% of experiment, and viscosities range from 2.6 to 3.8~cP. Onsager conductivities, which include correlated ion motion, are approximately 11.8, 8.5, and 9.8~mS\,cm$^{-1}$ for E1, E2, and E3. These values are lower than the corresponding Nernst--Einstein estimates and experimental conductivities, with the largest discrepancies from experiment in E2 and E3.

The mean solvent coordination numbers around Li$^+$ are 3.96, 3.98, and 4.00 for E1, E2, and E3, while the mean anion coordination numbers are 0.21, 0.23, and 0.18. Mean solvent coordination changes little across the three electrolytes and alone cannot explain their different conductivities. The reduction from Nernst--Einstein to Onsager conductivity reflects ion correlations. The finite-field response of the neutral solvent is approximately 18 for E1 and 22.5 for E2 and E3, consistent with E1's lower EC fraction.

\FloatBarrier
\subsection{Alloys}
\label{sec:cases-alloys}

\subsubsection{Composition Search for a Melting Target}
\label{sec:case-alloy-search}

\begin{figure}[!htb]
    \centering
    \includegraphics[width=\linewidth]{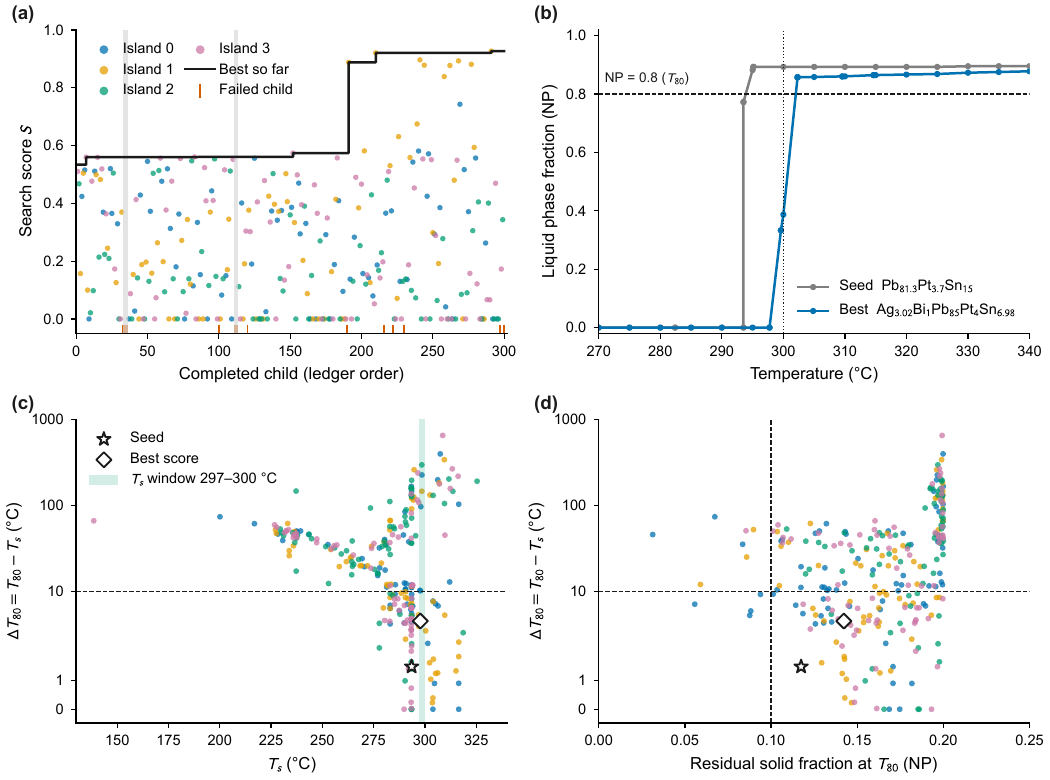}
    \small
    \caption{Composition search for a 300~°C melting target. (a) Search score of each valid child in order of completion, colored by island, with the best score so far. Ticks below the axis mark children without a result, and grey bands mark the two interrupted intervals. (b) Predicted liquid phase fraction near 300~°C for the seed and the highest-scoring composition (wt.\%). (c,d) Melting interval $\Delta T_{80}$ against $T_s$ and against the residual solid fraction at $T_{80}$ for 282 valid children. Dashed lines mark the final thresholds ($\Delta T_{80}=10$~°C, residual 0.1), and the green band marks the preferred $T_s$ window. Three children whose $T_{80}$ reached the scan limit are omitted from (c,d).}
    \label{fig:case-alloy-search}
\end{figure}

MatOS searches for alloys that melt near 300~°C over a narrow interval, a requirement for high-temperature joining and die attachment\cite{zeng2012hts}. A language model proposes compositions in a four-island evolutionary search\cite{romeraparedes2024funsearch}, and a composition-based surrogate predicts equilibrium phase fractions from 25 to 1000~°C at 1~bar. Each island starts from Pb$_{81.286}$Pt$_{3.714}$Sn$_{15}$ (wt.\%). Candidates contain two to six elements drawn from Ag, Au, Bi, Cd, Cu, Ge, In, Ni, Pb, Pd, Pt, Sb, Sn, and Zn, on a 0.01~wt.\% grid with Au below 20~wt.\% and Sn above 5~wt.\%. The language model selects a bounded composition region and proposes a composition within it. Pettifor-scale\cite{pettifor1984scale} and mat2vec\cite{tshitoyan2019unsupervised} representations identify near-duplicate proposals. From each predicted phase-fraction curve, the workflow extracts the solidus $T_s$, defined by a liquid fraction of $10^{-4}$ on cooling, and $T_{80}$, the lowest temperature at which the liquid fraction reaches 0.8. The score weights proximity of $(T_s+T_{80})/2$ to 300~°C, a narrow interval $\Delta T_{80}=T_{80}-T_s$, a solidus in the 297 to 300~°C window, and a low residual solid fraction at $T_{80}$ by 0.3, 0.3, 0.3, and 0.1, respectively. Four- to six-element compositions receive a 0.05 bonus. Elite retention, migration every 50 completed proposals, and summaries of previous results every 20 proposals guide subsequent search.

The search evaluates 286 distinct compositions, including the seed (Figure~\ref{fig:case-alloy-search}). The best score increases from 0.534 to 0.927 for Ag$_{3.02}$Bi$_{1}$Pb$_{85}$Pt$_{4}$Sn$_{6.98}$ (wt.\%), with predicted $T_s=297.75$~°C, $T_{80}=302.30$~°C, and $\Delta T_{80}=4.55$~°C. Its midpoint is within 0.03~°C of the target. The eight highest-scoring compositions belong to the same Pb--Ag--Pt--Sn family, some with approximately 1~wt.\% Bi. None meets all four acceptance criteria: $T_s>297$~°C, $\Delta T_{80}<10$~°C, residual solid fraction below 0.1, and a midpoint within 10~°C of 300~°C. Of the evaluated compositions, 56 satisfy the solidus criterion, 105 satisfy the interval criterion, and 11 satisfy the residual criterion. All 11 low-residual compositions have solidus temperatures at or below 293.5~°C, whereas the highest-scoring composition has a residual of 0.142. Within the sampled compositions, reducing residual solid competes with retaining a solidus near 300~°C. Thermodynamic calculations or measurements are needed to verify these surrogate predictions.

\FloatBarrier
\subsubsection{Melting-Temperature Prediction Model}
\label{sec:case-duocompnet}

The task is to develop a composition-based surrogate for predicting the melting-interval midpoint of Sn-containing alloys, $T_{\mathrm{mid}}=(T_{\mathrm{solidus}}+T_{\mathrm{liquidus}})/2$, in the 200 to 400~°C range. MatOS supports this development through an active-learning workflow that selects compositions for thermodynamic calculations and retrains the predictors as labels return. Within this workflow, we developed DuoCompNet and evaluated it alongside five baseline models.

\begin{figure}[!htb]
    \centering
    \includegraphics[width=\linewidth]{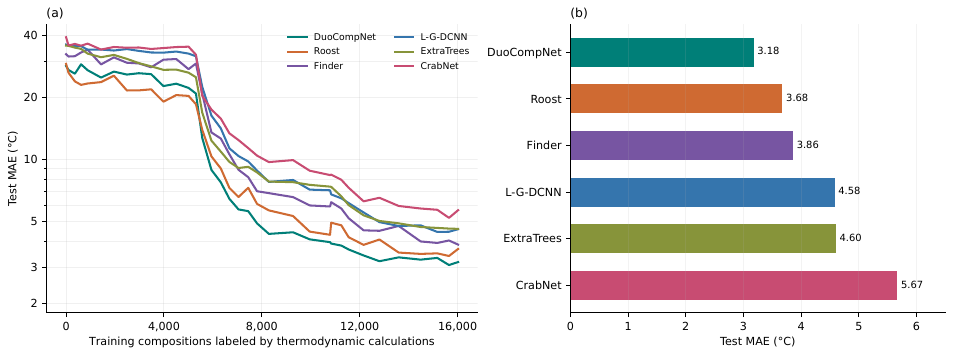}
    \small
    \caption{Test errors for six alloy melting-temperature models. (a) Learning curves over 35 training stages, plotted against the number of training compositions labeled by thermodynamic calculations, on a logarithmic MAE axis. (b) Final model comparison. Both use the same 1,401-composition test set of Sn-containing alloys with melting-interval midpoints from 200 to 400~°C. Reference values come from thermodynamic calculations with a 5~K temperature step.}
    \label{fig:case-duocompnet-curves}
\end{figure}

\begingroup
\clubpenalty=10000
\widowpenalty=10000
Across 35 training stages, we compare DuoCompNet with ExtraTrees\cite{geurts2006extratrees}, L-G-DCNN\cite{du2024lgdcnn}, Roost\cite{goodall2020roost}, CrabNet\cite{wang2021crabnet}, and Finder\cite{ihalage2022finder}, evaluated here using a continuous-fraction formula-graph implementation (Figure~\ref{fig:case-duocompnet-curves}). Each stage uses common training compositions, a separate 1,402-composition validation set, and a fixed 1,401-composition test set. DuoCompNet also uses available auxiliary thermodynamic labels during training. Neural models train for up to 300 epochs, with checkpoints selected on validation MAE. The final training set contains 22,396 compositions, including 16,020 entries labeled by thermodynamic calculations.
\par\endgroup

\begin{samepage}
As the dataset expands, DuoCompNet's test MAE falls from 28.40 to 3.18~°C, and Finder's from 32.37 to 3.86~°C. DuoCompNet has the lowest final MAE, followed by Roost (3.68~°C), Finder (3.86~°C), L-G-DCNN (4.58~°C), ExtraTrees (4.60~°C), and CrabNet (5.67~°C). Its error is 13.6\% lower than Roost's.\par
\end{samepage}

\begin{figure}[!htb]
    \centering
    \includegraphics[width=\linewidth]{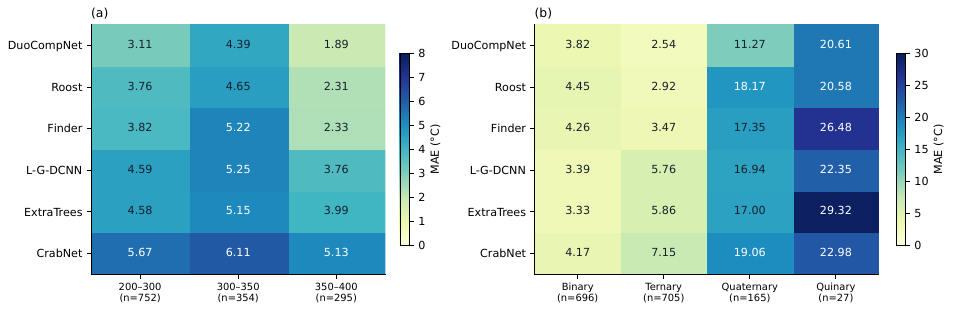}
    \small
    \caption{Dependence of test MAE on (a) melting-interval midpoint and (b) alloy component count. The binary and ternary compositions form the 1,401-composition test set. Separate quaternary and quinary test sets contain 165 and 27 compositions. All test compositions contain Sn and are excluded from training. Sample counts appear below the columns, and the heatmaps use separate color scales.}
    \label{fig:case-duocompnet-breakdown}
\end{figure}

DuoCompNet has the lowest test MAE in all three temperature intervals (Figure~\ref{fig:case-duocompnet-breakdown}), but model rankings vary with alloy composition. ExtraTrees leads on binary alloys at 3.33~°C, while DuoCompNet leads on ternaries at 2.54~°C. DuoCompNet's errors increase to 11.27~°C for quaternaries and 20.61~°C for quinaries, where the training set contains 607 and 159 compositions, respectively. After the final data addition, DuoCompNet's MAE on the 1,401-composition test set changes from 3.07 to 3.18~°C, while Finder's falls from 4.04 to 3.86~°C. The largest remaining errors occur in alloys with more components, making quaternary and quinary compositions a priority for further thermodynamic calculations.

\FloatBarrier

\section{Related Work}
\label{sec:related-work}

\subsection{Computational and Experimental Infrastructure for Materials Research}
Computational and experimental platforms automate different stages of materials research. FireWorks supports runtime changes to task graphs and failure recovery\cite{jain2015fireworks}. AiiDA uses data provenance and plugins to enable reproducible calculations and connections to external software\cite{huber2020aiida}. The atomate2 framework separates reusable workflow components from the execution engine to connect multiple simulation codes and machine-learning potentials\cite{ganose2025atomate2}. On the experimental side, a self-driving thin-film laboratory and a mobile robotic chemist select new experiments from measurement feedback\cite{macleod2020selfdriving,burger2020robotic,dai2024mobilerobots}. A-Lab runs robotic solid-state synthesis guided by active learning\cite{szymanski2023alab}, and CRESt combines multimodal information with robot operations to optimize electrocatalysts\cite{zhang2025crest}. These platforms provide dynamic scheduling, plugin extension, error recovery, or experimental feedback (Appendix~\ref{app:materials-background} gives further background). MatOS reuses existing solvers and workflow engines, and represents methods, checking procedures, and process experience as reusable scientific assets that computational and experimental stages can share.

\subsection{Research Agents for Materials and Chemistry}
Research agents for materials and chemistry differ mainly in how they organize capabilities. Early systems connect language models to scientific software and materials databases for synthesis planning, property prediction, and knowledge retrieval. ChemCrow uses 18 chemistry tools for planning and experimental tasks\cite{bran2024chemcrow}, CACTUS couples a language model with cheminformatics tools\cite{mcnaughton2024cactus}, ChatMOF combines an agent, toolbox, and evaluator for metal--organic framework tasks\cite{kang2024chatmof}, and LLaMP uses hierarchical agents to connect information sources such as Materials Project with atomistic simulation tools\cite{chiang2025llamp}. Honeycomb combines a materials knowledge base with a tool hub\cite{zhang2024honeycomb}. In computational settings, VASPilot wraps VASP operations with the Model Context Protocol (MCP)\cite{anthropic2024mcp}, restarts failed calculations after adjusting parameters from error messages, and proposes extensions to other DFT codes\cite{liu2025vaspilot}. ChatMat coordinates retrieval, DFT, and machine-learning models through specialized roles and permits additional agents\cite{lv2026chatmat}. SciToolAgent schedules hundreds of scientific tools through a tool knowledge graph\cite{ding2025scitoolagent}. Several multi-agent systems target computational materials science, including modular agents for multiple tasks\cite{chaudhari2026matsciagent}, planner-driven materials discovery\cite{wang2025matagent}, physics-aware reasoning for inorganic materials\cite{ghafarollahi2026sparksmatter}, and collaborative lightweight models for crystal materials\cite{shi2026matbrain}. Other systems couple agents with automated laboratories. In chemistry, Coscientist by Boiko \emph{et al.} plans and executes experiments autonomously\cite{boiko2023coscientist}. MARS integrates 19 agents with robotic experiments on perovskite nanocrystals\cite{shi2026mars}, ChemAgents runs design--synthesis--characterization--optimization loops for high-entropy catalysts\cite{song2025chemagents}, and the collective-intelligence system of Li \emph{et al.} uses specialized models to propose synthesis plans that are then tested experimentally\cite{li2026collective}. MatClaw combines code execution with hierarchical memory for materials exploration\cite{zhang2026matclaw}, and the MIRA technical report describes versioned scientific skills, a persistent knowledge wiki, and computational and experimental platforms\cite{mira2026general}. Beyond a single domain, the general-purpose Co-Scientist system uses multiple agents to generate and refine research hypotheses\cite{gottweis2026coscientist}. MatOS represents data, knowledge, tools, skills, roles, and workflows as nodes of a single graph with typed relations, and uses records from long runs to revise several related assets together, with each revision checked before adoption.

\subsection{Agent Harnesses and Reliability of Long-Horizon Tasks}
General agent harnesses support multiagent collaboration, execution isolation, and state management for multistep work. AutoGen supports multiagent conversation and tool coordination\cite{wu2024autogen}, OpenHands provides an event-driven execution environment and SDK for software agents\cite{wang2025openhands,wang2026openhandsSDK}, and LangGraph supports recovery, state branching, and cross-session memory through checkpoints and persistent storage\cite{langchain2026langgraphpersistence}. For coding, Claude Code supports skills, subagents, and saveable dynamic workflows\cite{anthropic2025claudecode,anthropic2026claudeworkflows}. Codex spans CLI, IDE, and cloud tasks\cite{openai2025codex}. OpenCode provides configurable agents, skills, permissions, and plugins\cite{sst2026opencode}. DeepSeek Harness (dsh) represents models, tools, skills, sessions, and scheduling as composable plugins and uses an append-only event log for recovery and replay\cite{deepseekai2026harness}. These harnesses provide extensibility, persistence, and dynamic orchestration. Reliability, however, has improved more slowly than accuracy: evaluating consistency across runs, robustness, predictability, and safety, Rabanser \emph{et al.} found that recent capability gains yielded only small improvements in reliability\cite{rabanser2026reliability}. Materials research adds heterogeneous simulation codes, long computations, and experimental feedback, and its conclusions depend on material structures, models, and parameter choices\cite{lejaeghere2016repro}. After a plan change or interruption, the system must examine existing computational inputs, process records, and artifacts to decide whether they still support subsequent analysis\cite{huber2020aiida}. MatOS ties the execution graph to the scientific assets it uses, so that methods and validation evidence are kept when a long-running job is resumed.

\subsection{Agent Experience Learning and Self-Evolution}
An agent can improve after deployment by updating different objects. Reflexion adjusts later behavior through verbalized feedback, ExpeL extracts cross-task experience from trajectories, and Voyager accumulates executable skills\cite{shinn2023reflexion,zhao2024expel,wang2024voyager}. Agent Workflow Memory and AutoManual summarize reusable processes and rules, respectively\cite{wang2025awm,chen2024automanual}. Prompts, system structure, and tool-use strategies can also change: PromptBreeder, ADAS, and AFlow explore prompt, agent-design, and workflow search\cite{fernando2023promptbreeder,hu2024adas,zhang2025aflow}. ToolEVO addresses changes to tool interfaces\cite{chen2024toolevo}. G\"{o}del Agent and Darwin G\"{o}del Machine study code or architecture modification and its evaluation\cite{yin2025godel,dgm2025}. A survey organizes the field by what evolves, when, and how\cite{gao2026selfevolving}. For scientific tools, SciToolAgent-Evo uses skills, experience, and a tool ontology graph to acquire tools in open environments\cite{tang2026scitoolagentevo}. These works update verbal experience, skill libraries, tool knowledge, or agent designs, and most of them were evaluated on general-domain benchmarks such as question answering, coding, web navigation, and games. MatOS uses task trajectories and user feedback to identify methods, tools, and validation procedures that need coordinated revision, and structural checks and semantic review assess the validity and applicability of each revision.

\subsection{Agent Operating Systems and Scientific Asset Management}
Agent OS research brings resource management, versioning, and controlled updates to agent systems. AIOS schedules language-model agents and manages resources\cite{mei2024aios}. Cerebrum offers a development and distribution platform with models, memory, tools, and an Agent Hub\cite{rama2025cerebrum}. Autogenesis models prompts, agents, tools, environments, and memory as resources with lifecycle and version interfaces and uses auditable operators for updates and rollback\cite{zhang2026autogenesis}. Other public prototypes explore microkernels, container execution, and capability security\cite{xkernel,substrate,jordan2025agentos}. In materials research, these mechanisms also need to record where a method applies, which physical checks it requires, and which computational or experimental evidence supports each conclusion. MatOS applies them to scientific assets, covering their composition, execution, and revision.

Table~\ref{tab:rw-comparison} compares representative systems in asset organization, stateful execution, and experience-driven evolution, and indicates whether each has a materials-specific design.

\begin{table}[!htb]
\centering
\footnotesize
\newcommand{\rwdash}{{\color{black!28}\leaders\hbox to 3.5pt{\hss\rule[0.6ex]{2pt}{0.4pt}\hss}\hfill}}
\setlength{\tabcolsep}{4pt}
\renewcommand{\arraystretch}{1.16}
\caption{Capabilities of representative systems, grouped as in Section~\ref{sec:related-work}. \cmark\ indicates an explicit mechanism. \xmark\ indicates that no explicit mechanism was identified.}
\label{tab:rw-comparison}
\begin{tabular}{@{}>{\raggedright\arraybackslash}p{3.3cm}
  >{\centering\arraybackslash}m{1.4cm}
  >{\centering\arraybackslash}m{1.6cm}
  >{\centering\arraybackslash}m{1.4cm}
  >{\centering\arraybackslash}m{1.6cm}
  >{\centering\arraybackslash}m{1.7cm}
  >{\centering\arraybackslash}m{1.4cm}
  >{\centering\arraybackslash}m{1.5cm}@{}}
\toprule
 & \multicolumn{2}{c}{\textbf{Asset organization}} & \multicolumn{2}{c}{\textbf{Stateful execution}} & \multicolumn{2}{c}{\textbf{Self-evolution}} & \textbf{Domain} \\
\cmidrule(lr){2-3}\cmidrule(lr){4-5}\cmidrule(l){6-7}
\textbf{System} & \textbf{Modular extension} & \textbf{Capability links} & \textbf{Dynamic orchestration} & \textbf{Persistent recovery} & \textbf{Experience extraction} & \textbf{Update acceptance} & \textbf{Materials focus} \\
\midrule
\addlinespace[1pt]
\multicolumn{8}{c}{\textit{Scientific workflow systems}} \\[-1.1ex]
\multicolumn{8}{@{}c@{}}{\rwdash} \\[-0.8ex]
AiiDA\cite{huber2020aiida} & \cmark & \cmark & \cmark & \cmark & \xmark & \xmark & \cmark \\
FireWorks\cite{jain2015fireworks} & \cmark & \xmark & \cmark & \cmark & \xmark & \xmark & \xmark \\
\midrule
\multicolumn{8}{c}{\textit{Scientific research agents}} \\[-1.1ex]
\multicolumn{8}{@{}c@{}}{\rwdash} \\[-0.8ex]
ChemCrow\cite{bran2024chemcrow} & \cmark & \xmark & \cmark & \xmark & \xmark & \xmark & \xmark \\
MIRA\cite{mira2026general} & \cmark & \cmark & \cmark & \xmark & \cmark & \xmark & \cmark \\
SciToolAgent-Evo\cite{tang2026scitoolagentevo} & \cmark & \cmark & \cmark & \xmark & \cmark & \cmark & \cmark \\
\midrule
\multicolumn{8}{c}{\textit{General agent frameworks and coding agents}} \\[-1.1ex]
\multicolumn{8}{@{}c@{}}{\rwdash} \\[-0.8ex]
LangGraph\cite{langchain2026langgraphpersistence,langchain2026langgraphgraphapi} & \cmark & \xmark & \cmark & \cmark & \xmark & \xmark & \xmark \\
Claude Code\cite{anthropic2026claudeworkflows,anthropic2026claudememory} & \cmark & \cmark & \cmark & \cmark & \cmark & \xmark & \xmark \\
Codex CLI\cite{openai2026codexcli,openai2026codexmemorypipeline} & \cmark & \cmark & \cmark & \cmark & \cmark & \xmark & \xmark \\
OpenCode\cite{sst2026opencode} & \cmark & \cmark & \cmark & \cmark & \xmark & \xmark & \xmark \\
DeepSeek Harness\cite{deepseekai2026harness} & \cmark & \cmark & \cmark & \cmark & \xmark & \xmark & \xmark \\
\midrule
\multicolumn{8}{c}{\textit{Agent operating systems}} \\[-1.1ex]
\multicolumn{8}{@{}c@{}}{\rwdash} \\[-0.8ex]
AIOS\cite{mei2024aios} & \cmark & \xmark & \cmark & \cmark & \xmark & \xmark & \xmark \\
Autogenesis\cite{zhang2026autogenesis} & \cmark & \cmark & \cmark & \xmark & \cmark & \cmark & \xmark \\
\midrule
\textbf{MatOS (this work)} & \cmark & \cmark & \cmark & \cmark & \cmark & \cmark & \cmark \\
\bottomrule
\end{tabular}
\end{table}

\section{Conclusion}
We introduced MatOS, an agent operating system for materials research. MatOS organizes data, knowledge, tools, skills, roles, and workflows as reusable scientific assets in a layered graph, so that new capabilities can be added by registering assets and combined for new tasks. A stateful adaptive workflow carries out each task as an explicit execution graph with scientific checks and keeps the records needed to pause, resume, and trace long-running work. Evidence-driven asset evolution turns these records and user feedback into checked revisions of the assets, so that experience gained in one task benefits later ones.

With the same base model, MatOS improves on the previously best-performing agent on ResearchClawBench by 11.15\%, and its DeepSeek-V4-Flash configuration also outperforms that agent at a comparable cost. With DeepSeek-V4-Pro as the common base model, it solves more MatTools problems and SciAgentGym cases than Codex CLI and Claude Code. In eight research cases, MatOS narrowed more than 30 million candidate structures to 97 solid-electrolyte candidates, identified 89 halide variants with higher computed conductivity than their parent structures, linked the force errors of 11 pretrained interatomic potentials to local atomic environments, and supported a training-data selection method and an alloy melting-temperature model that outperform their baselines. It also reproduced the CFDBench lid-driven cavity in agreement with the classical reference solution and found inconsistent metadata in the benchmark's geometry subset.

\begingroup
\clubpenalty=10000
\widowpenalty=10000
These results show that organizing research capabilities as reusable, evolving assets enables agents to carry out long, multistage materials studies, from screening tens of millions of structures to developing new methods. Because every new database, code, or method enters MatOS as an asset that later tasks can reuse, and every completed task can improve the assets it used, the capabilities available to agents accumulate with use rather than being rebuilt for each study. The same assets and workflow runtime extend directly to instruments and experimental protocols, allowing a single study to link computational and experimental evidence.
\par\endgroup

\phantomsection
\addcontentsline{toc}{section}{Contributors}
\section*{Contributors}
\label{sec:contributors}

\textbf{Core contributors.} Dingyang Lyu\textsuperscript{1,2,3,\dag} (\email{s-ldy25@bza.edu.cn}) designed the architecture of MatOS, developed its core asset, workflow, and evolution mechanisms, implemented most of the system, and drafted the initial manuscript of this technical report. Hongwei Du\textsuperscript{1,2} provided scientific direction for the project, including defining the scope of scientific assets and planning the scientific case studies. He carried out six of the eight research cases (Sections~\ref{sec:cases-potentials}--\ref{sec:case-liquid-electrolyte} and~\ref{sec:case-duocompnet}) and led the English-language revision and formatting of the manuscript. Baole Wei\textsuperscript{1,2} established the project and guided key algorithm design and experimental planning. He also contributed to English polishing, logical refinement, and project coordination.

\textbf{Contributors.} Yongheng Li\textsuperscript{1,2} performed the alloy composition search study (Section~\ref{sec:case-alloy-search}) and reviewed the manuscript. Jing Gao\textsuperscript{1,2,\dag} contributed the knowledge assets of MatOS and reviewed the manuscript. Zhonghao Chang\textsuperscript{2,4,\dag} implemented part of the code for evidence-driven asset evolution. Jingjie Huang\textsuperscript{2,4,\dag} contributed to the CFDBench reproduction case (Appendix~\ref{app:cfdbench-cavity}). Ziheng Lu\textsuperscript{1,2,5} provided overall guidance for the project. Xiangran Cheng\textsuperscript{1,2}, Xin He\textsuperscript{1,2}, Yongda Huang\textsuperscript{1,2}, Zhen Lei\textsuperscript{3}, Zefeng Li\textsuperscript{1,2}, Jiahui Liu\textsuperscript{1,2}, Jian Wen\textsuperscript{1,2}, Feng Yu\textsuperscript{1,2}, and Bonan Zhu\textsuperscript{1,2,6} (listed alphabetically by surname) assisted with manuscript preparation.

\textbf{Project leads.} Baole Wei and Hongwei Du led and managed the project.

\textbf{Corresponding authors.} Ziheng Lu (\email{luziheng@zgci.ac.cn}), Baole Wei (\email{weibaole@zgci.ac.cn}), Hongwei Du (\email{duhongwei@zgci.ac.cn}), and Yongheng Li (\email{liyongheng@zgci.ac.cn}).

\medskip
\noindent\rule{0.3\linewidth}{0.4pt}\par
\noindent{\small
\textsuperscript{1}Zhongguancun Academy, Beijing, China\\
\textsuperscript{2}Zhongguancun Institute of Artificial Intelligence, Beijing, China\\
\textsuperscript{3}Institute of Automation, Chinese Academy of Sciences, Beijing, China\\
\textsuperscript{4}Tsinghua University, Beijing, China\\
\textsuperscript{5}Kairos Materials\\
\textsuperscript{6}Beijing Institute of Technology, Beijing, China\\
\textsuperscript{\dag}Work carried out during the research program at Zhongguancun Academy and Zhongguancun Institute of Artificial Intelligence.\par}

\clearpage
\phantomsection
\addcontentsline{toc}{section}{\refname}
\bibliographystyle{unsrt}
\bibliography{main}
\clearpage
\phantomsection
\addcontentsline{toc}{section}{Appendix}
\beginappendix
\section{Composition and Coverage of Scientific Assets}
\label{app:asset-coverage}

\begin{figure}[!htb]
    \centering
    \includegraphics[width=\linewidth]{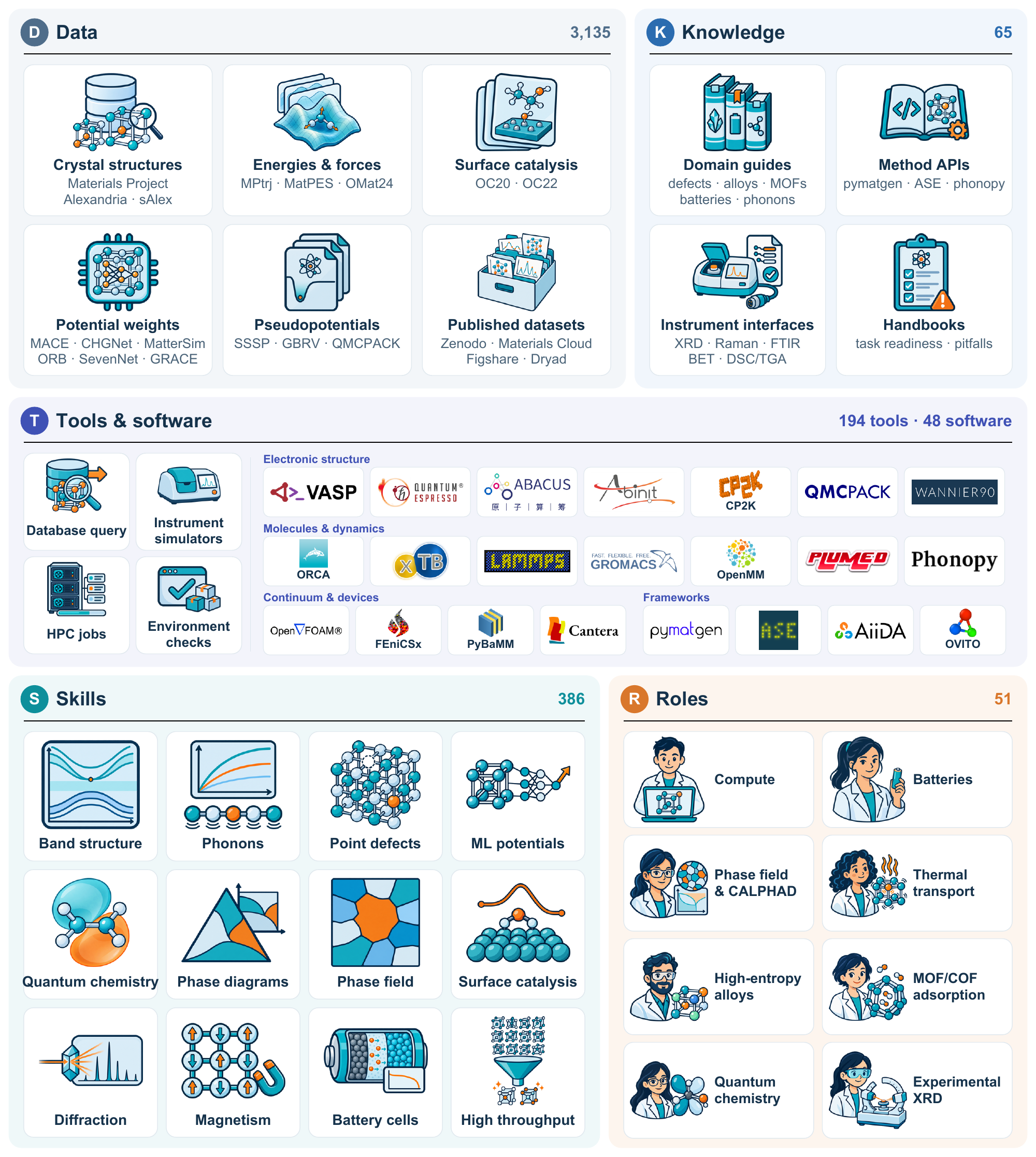}
    \caption{Composition and coverage of the MatOS scientific asset library. The five regions show data ($D$), knowledge ($K$), tools and specialized software ($T$), skills ($S$), and roles ($R$), with asset counts at the top right. The data region lists representative datasets, pretrained potentials, and pseudopotential libraries by research object. The knowledge region shows domain guides, method-interface knowledge, instrument interface specifications, and handbooks. In the tool region, callable research interfaces appear on the left and representative integrated software on the right. The skill and role regions list the main method families and specialized roles. Illustrations are schematic. Software marks belong to their respective owners.}
    \label{fig:asset-coverage}
\end{figure}

The scientific asset library spans electronic structure, lattice dynamics and thermal transport, defects and surfaces, machine-learning interatomic potentials, molecular quantum chemistry, phase diagrams and microstructure, sorption in porous materials, battery electrochemistry, and continuum computation (Figure~\ref{fig:asset-coverage}, Table~\ref{tab:asset-coverage-examples}). Assets supply content through documents, implementations, or external resource references and connect through the relations defined in Section~\ref{sec:asset-graph}. Figure~\ref{fig:asset-workflows} shows four representative execution graphs.

\begin{table}[!htb]
    \centering
    \small
    \renewcommand{\arraystretch}{1.22}
    \caption{Scale and representative content of MatOS scientific assets. Tool assets provide callable interfaces to solver backends, including 48 integrated materials software packages.}
    \label{tab:asset-coverage-examples}
    \begin{tabular}{@{}p{0.13\linewidth}r p{0.72\linewidth}@{}}
        \toprule
        \textbf{Type} & \textbf{Count} & \textbf{Representative content} \\
        \midrule
        Data $D$ & 3,135 & Structure databases (Materials Project and Alexandria), training data (MPtrj, MatPES, OMat24, and OC20/OC22), pretrained potentials (MACE, CHGNet, and MatterSim), SSSP and GBRV pseudopotentials, and published research datasets \\
        Knowledge $K$ & 65 & Domain guides and code-resource assessments for twelve research areas, pymatgen method-interface knowledge, instrument interface specifications, indexes of datasets and the machine-learning-potential ecosystem, and handbooks on task readiness and common pitfalls \\
        Tool $T$ & 194 & Database and literature retrieval. Catalogs of models, pseudopotentials, and computing environments. Simulators for XRD, Raman, FTIR, BET, and thermal analysis. High-performance computing job submission and monitoring. PyBaMM solution and Battery Parameter eXchange (BPX) validation \\
        Skill $S$ & 386 & Procedures for band structure and electronic structure, phonons and thermal transport, charged defects, machine-learning potentials, high-throughput screening, molecular quantum chemistry, phase diagrams and phase fields, catalysis and electrochemistry, spectroscopy, and magnetism \\
        Role $R$ & 51 & A computation coordinator and subroles for electronic structure, molecules, phase stability, and other calculations. Domain roles for batteries, phase fields and CALPHAD, thermal transport, high-entropy alloys, porous-material sorption, and experimental XRD. Roles for workflow planning, review, and self-evolution \\
        Workflow $W$ & 104 & Stability screening, gas-sorption screening, charged-defect formation energies, Quantum ESPRESSO (QE) and AiiDA calculation chains, machine-learning-potential benchmarks, diffraction phase analysis, battery discharge simulation, and research and asset-construction planning \\
        \bottomrule
    \end{tabular}
\end{table}

\subsection{Research Coverage of the Six Asset Types}

\paragraph{Data assets ($D$).}
The 3,135 datasets provide inputs and references for modeling, training, screening, and result comparison. Structure and property databases supply starting points for crystal modeling. Materials Project provides crystal structures and properties computed with consistent settings\cite{jain2013materialsproject}, and Alexandria\cite{schmidt2024alexandria} contains about five million relaxed inorganic structures computed with PBE, PBEsol, and SCAN\cite{perdew1996pbe,perdew2008pbesol,sun2015scan}. Its sAlex subset, filtered against the WBM benchmark set, is registered separately\cite{barrosoluque2026omat24,wang2021wbm}. Potential-energy-surface data support the training and evaluation of machine-learning potentials. MPtrj\cite{deng2023chgnet} contains about 1.58 million structures from 146,000 compounds. MatPES provides PBE and r$^2$SCAN\cite{furness2020r2scan} reference sets\cite{kaplan2025matpes}, while OMat24\cite{barrosoluque2026omat24} contains about 118 million inorganic single-point calculations. OC20/OC22\cite{chanussot2021oc20,tran2023oc22} cover surface catalysis. Pretrained potentials are registered as data assets that record model weights and their provenance. MACE-MP-0\cite{batatia2025macemp} is trained on MPtrj and covers 89 elements, while MACE-OFF23 targets organic molecules\cite{kovacs2025maceoff}. Other registered universal potentials include CHGNet\cite{deng2023chgnet}, M3GNet\cite{chen2022m3gnet}, MatterSim\cite{yang2024mattersim}, ORB\cite{neumann2024orb}, SevenNet\cite{park2024sevennet}, and GRACE\cite{bochkarev2024grace}. The SSSP\cite{prandini2018sssp} and GBRV\cite{garrity2014gbrv} pseudopotentials and the QMCPACK pseudopotential library\cite{pseudolib} needed for first-principles calculations, together with the BPX format\cite{bpx}, are also managed as data assets. In addition, the library registers published research data from repositories such as Zenodo, Materials Cloud\cite{talirz2020materialscloud}, Figshare, and Dryad as dataset-level or paper-level entries. Each data asset records its provider, license, access route, and local mount location. Large datasets are registered by reference and retrieved to compute nodes when a task requires them, and their dependencies point to the corresponding retrieval tools and background knowledge.

\paragraph{Knowledge assets ($K$).}
The 65 knowledge assets record the physical basis, applicability conditions, and selection criteria of methods. Domain guides cover twelve research areas: defects and surfaces, crystal structure prediction, high-entropy alloys, sorption in porous frameworks, phonons and thermal transport, electronic transport, battery interfaces, catalytic microkinetics, reaction networks, amorphous materials and liquids, phase-field microstructure, and active learning of machine-learning potentials. For each area, the guides also assess the purpose and maturity of the available code repositories. The defect guide, for instance, treats pydefect\cite{kumagai2021pydefect} and pymatgen-analysis-defects\cite{shen2024pmgdefects} as references for defect objects and workflow concepts, lists ShakeNBreak\cite{mosqueralois2022snb} as a candidate workflow for searching reconstructed defects, and records which of them can be run in the deployed environment. Method-interface knowledge centers on pymatgen and explains the correct use of more than twenty interface groups, including structure and symmetry, phase diagrams and thermodynamic stability, elastic tensors, band structures and densities of states, surfaces and Wulff shapes, XRD pattern calculation, and interoperation with ASE and phonopy. Instrument interface specifications define a common calling interface for simulated characterization instruments and laboratory equipment. Indexes of datasets and of the machine-learning-potential ecosystem record the scale and license of each dataset and the data asset that represents it. A task-readiness handbook and a checklist of common pitfalls record recurring methodological errors as searchable criteria, which agents consult to settle modeling assumptions before calling tools.

\paragraph{Tool assets ($T$).}
The 194 tool assets wrap retrieval, solution, validation, and job management as callable interfaces with explicit inputs and outputs. Data and literature tools support online and local Materials Project queries, paper retrieval, and document parsing. Catalog and preflight tools report the availability of machine-learning-potential weights, QE pseudopotentials, software licenses, executables, Conda computing environments, and the atomistic, battery, and combustion toolchains, and the associated skills and roles stop a task when a required item is missing. Instrument tools provide simulators for XRD, Raman, FTIR, UV-Vis, BET surface area, DSC/TGA thermal analysis, and four-point-probe conductivity, together with DARA-based XRD phase analysis\cite{fei2026dara}. Job tools submit, query, and cancel tasks on high-performance computing platforms and collect their logs and outputs. Dedicated solver tools run PyBaMM\cite{sulzer2021pybamm} cell models directly and validate BPX parameter files. At the software level, the system integrates 48 materials software packages. They span electronic structure (VASP, Quantum ESPRESSO, ABACUS, ABINIT, CP2K, Wannier90, and QMCPACK)\cite{kresse1996vasp,giannozzi2009qe,li2016abacus,gonze2020abinit,kuhne2020cp2k,pizzi2020wannier90,kim2018qmcpack}, molecular computation (ORCA, xtb, and CREST)\cite{neese2020orca,bannwarth2021xtb,pracht2024crest}, atomistic and molecular dynamics (LAMMPS, GROMACS, OpenMM, and PLUMED)\cite{thompson2022lammps,abraham2015gromacs,eastman2024openmm,tribello2014plumed}, lattice dynamics (phonopy)\cite{togo2015phonon}, continuum and device modeling (OpenFOAM\textsuperscript{\textregistered}, FEniCSx, PyBaMM, and Cantera)\cite{weller1998openfoam,baratta2023dolfinx,sulzer2021pybamm,goodwin2023cantera}, and materials computing frameworks such as pymatgen, ASE, AiiDA, atomate2, and OVITO\cite{ong2013pymatgen,larsen2017ase,huber2020aiida,ganose2025atomate2,stukowski2010ovito}.

\paragraph{Skill assets ($S$).}
The 386 scientific skills encode materials computation methods as reusable procedures. A skill usually has five parts: when to use it, how to choose a method, prerequisites, execution steps, and result checks. The skills fall into more than fifty method families. The electronic-structure family includes band structures, advanced band analysis, DFT corrections, Wannier functions, Fermi surfaces, and topological properties. The lattice-dynamics family includes phonons, electron--phonon coupling, and thermal conductivity. Other families cover defects and reactions, structure modeling, machine-learning potentials, high-throughput computation, molecular quantum chemistry, biomolecular dynamics, bonding analysis, Monte Carlo methods, phase transitions and phase diagrams, catalysis and electrochemistry, optics, ferroelectricity and piezoelectricity, mechanics, magnetism, two-dimensional materials, and battery electrodes. Skills also state the grounds for choosing between methods. The band-structure skill notes that a machine-learning potential has no electronic degrees of freedom and can only pre-relax the structure. Band structures are calculated by solving the Kohn--Sham equations with QE. The phonon skill chooses between finite displacements with MACE and phonopy and density-functional perturbation theory\cite{baroni2001dfpt} in QE according to the required accuracy and the cell size, and uses imaginary frequencies to judge dynamical stability\cite{togo2015phonon}. The defect skill covers the calculation of charged-defect formation energies, finite-size corrections, and charge-transition levels\cite{ganose2025atomate2}.

\paragraph{Role assets ($R$).}
The 51 roles combine assets for specific research responsibilities. Each has a responsibility prompt and linked skills, tools, knowledge, and data. The coordinating role \texttt{compute} routes density-functional, molecular-dynamics, and machine-learning-potential tasks and can delegate to computing subroles for electronic structure, molecules, phase stability, structures, defects and surfaces, and mechanics and thermal properties. Domain roles cover battery electrochemistry, phase fields and CALPHAD, phonons and thermal transport, high-entropy alloys, sorption in metal--organic frameworks (MOFs), covalent organic frameworks (COFs), and other porous materials, crystal structure prediction, amorphous materials and glasses, catalytic microkinetics, reaction networks, molecular quantum chemistry and molecular spectroscopy, and experimental XRD. For example, \texttt{battery-electrochem-modeler} mounts battery-electrochemistry and electrode skills, the PyBaMM solver and BPX validation tools, knowledge of macroscale battery modeling, and the BPX parameter specification. It carries out single-particle model (SPM), SPM with electrolyte (SPMe)\cite{marquis2019spme}, and Doyle--Fuller--Newman (DFN)\cite{doyle1993dfn} cell simulations and checks parameter portability, and it can hand atomistic questions about electrodes and interfaces to the defect and surface computing role. Other roles serve the system itself. Roles for workflow planning, building, review, repair, and optimization design and maintain execution graphs, while roles for user-feedback critique, feedback classification, semantic acceptance, and experience refinement carry out the evidence critique and dual acceptance described in Section~\ref{sec:asset-evolution}.

\paragraph{Workflow assets ($W$).}
The 104 workflows express research goals as execution graphs with typed nodes, dependencies, and declared output artifacts. By purpose they fall into four groups. The first contains directly executable research processes, such as stability screening of known structures, gas-sorption screening in porous frameworks, high-throughput virtual screening of drug molecules, diffraction phase analysis, and battery discharge simulation. The second contains calculation-evidence workflows with QE as the backend, such as single and batched calculations, NEB, spin--orbit band structures, and charged-defect formation energies, together with wrapped AiiDA work chains for structure relaxation, equations of state, band structures, NEB, and convergence tests\cite{huber2020aiida}. The third contains benchmarks of machine-learning potentials, which fit equilibrium lattice constants from equation-of-state scans and compare them with experimental references. The fourth contains planning workflows. Some of these produce a checkpointed research plan before compute is committed, for example for halide solid-electrolyte discovery or battery-model parameter fitting. Others address the twelve research areas and convert domain guides and code-resource assessments into draft assets for review, in several cases after a short test run in the target domain. Across all four groups, agent nodes handle planning and interpretation, execution nodes run deterministic computation scripts, guard nodes, shown as check steps in Figure~\ref{fig:asset-workflows}, check intermediate products before the process advances, and subworkflow nodes reuse existing computational procedures.

\subsection{Representative Workflows}

\begin{figure}[!htb]
    \centering
    \includegraphics[width=\linewidth]{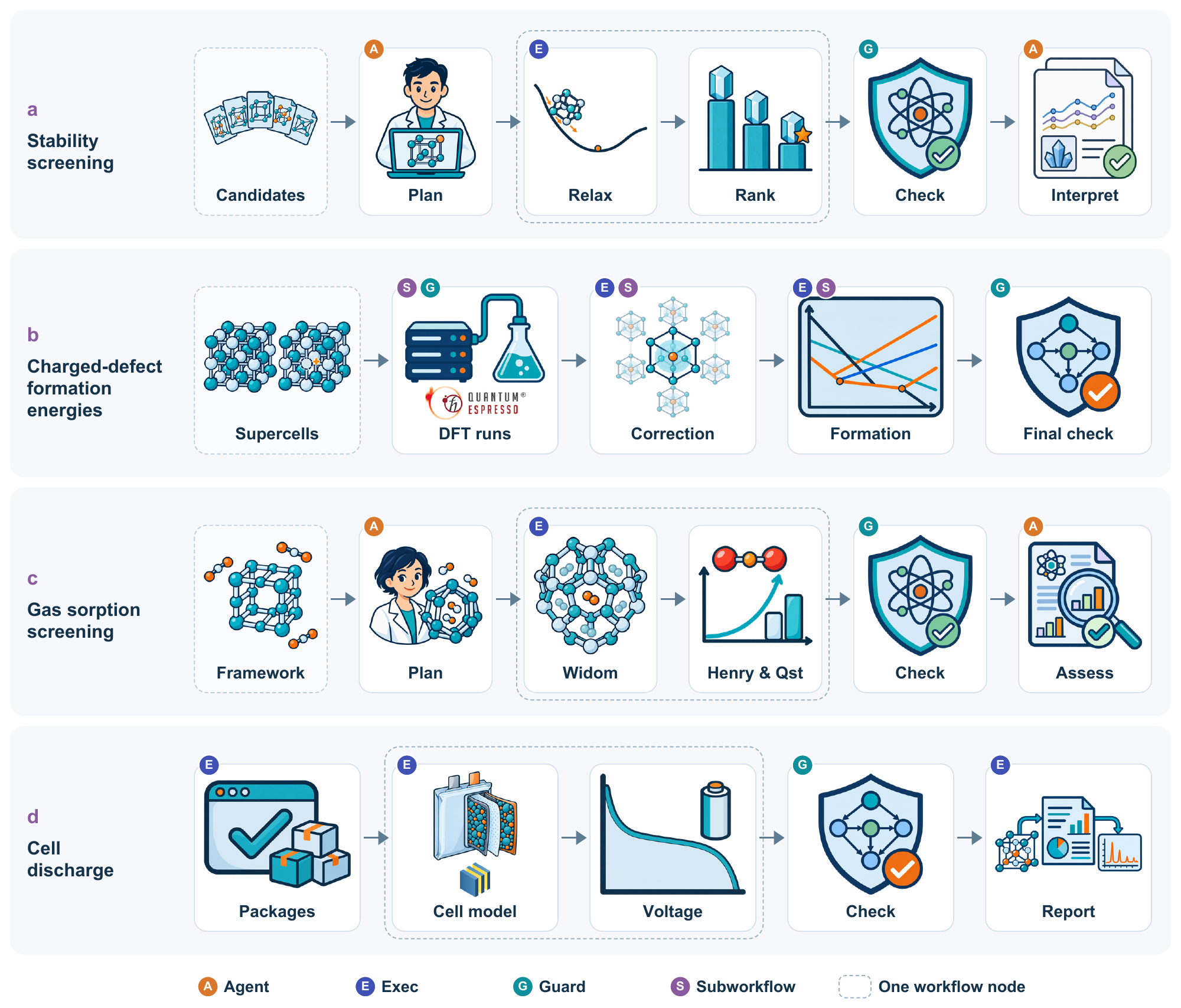}
    \caption{Execution structure of four representative workflows. (a) Stability screening of known candidate structures. (b) Charged-defect formation energies with QE as the computational backend. (c) Gas-sorption screening in porous frameworks. (d) PyBaMM cell-discharge simulation. Circular markers identify agent (A), execution (E), guard (G), and subworkflow (S) nodes. Steps inside a dashed box are carried out by a single execution node, and the leftmost dashed card is the input. Illustrations are schematic.}
    \label{fig:asset-workflows}
\end{figure}

Figure~\ref{fig:asset-workflows} shows the node structure of four representative workflows.

\paragraph{Stability screening.}
\texttt{atomistic-materials-discovery} relaxes a set of known candidate structures and ranks their stability (Figure~\ref{fig:asset-workflows}a). An agent node acting as the \texttt{compute} role states the screening objective, the candidate families, and the choice of calculator. An execution node reads up to 64 CIF structures of at most 512 atoms each, calls the effective-medium-theory (EMT) calculator\cite{jacobsen1996emt} or a local MACE model through ASE, relaxes each structure with the BFGS optimizer to a force threshold of 0.02~eV\,\AA$^{-1}$ within at most 200 steps, ranks the candidates by a cohesive-energy proxy, and writes a ranking table, a report, and a ranking plot. A guard node confirms that the ranking is valid and contains at least one successfully relaxed candidate. A final agent node interprets the ranking, identifying the most stable candidates and structural preferences, and proposes next steps such as DFT verification of the top-ranked structures. If the potential itself needs validation first, the machine-learning-potential benchmark workflow is run beforehand.

\paragraph{Charged-defect formation energies.}
\texttt{qe-defect-formation-suite} organizes several QE calculations as subworkflows (Figure~\ref{fig:asset-workflows}b)\cite{giannozzi2009qe}. The workflow first runs the pristine-supercell calculation and checks its scientific validity, then runs the supercell calculations for all defect states as a batch and validates each of them. An execution node next assembles the correction inputs and calls the charged-defect correction subworkflow, which computes finite-size corrections from potential-alignment and dielectric data. After an execution node assembles the energy ledger, the formation-energy subworkflow computes the formation energy of each defect state from the composition changes, chemical potentials, and electronic reference range. A final guard node and artifact validation fail the workflow unless every declared defect state has a compatible, completed calculation. The formation specification lists defect states and thermodynamic references. Total energies come from upstream calculation outputs.

\paragraph{Gas-sorption screening.}
\texttt{atomistic-sorption-discovery} estimates the low-loading sorption of an adsorbate in a porous framework (Figure~\ref{fig:asset-workflows}c). An agent node acting as the \texttt{porous-materials-sorption} role confirms the computing environment and states the screening plan. An execution node performs Widom test-particle insertion\cite{widom1963} in the Henry regime, with a single-site Lennard-Jones adsorbate, Universal Force Field (UFF)\cite{rappe1992uff} parameters for framework atoms, and Lorentz--Berthelot mixing rules. By default it makes 5,000 insertions at 298~K with a 12~\AA\ cutoff and outputs the Henry coefficient, the mean interaction energy, and the isosteric heat $Q_{\mathrm{st}}$. A guard node requires $Q_{\mathrm{st}}$ to be finite and the Henry factor to be positive. A final agent node assesses the binding strength and low-pressure uptake from these values. Both the adsorbate parameters and the framework CIF can be replaced, so the same workflow compares different gases, such as CO$_2$, N$_2$, and CH$_4$, and different frameworks. The result estimates sorption at infinite dilution. Full isotherms require grand-canonical Monte Carlo calculations.

\paragraph{Cell-discharge simulation.}
\texttt{battery-pybamm-cell-discharge} runs a lithium-ion cell discharge calculation at the device scale (Figure~\ref{fig:asset-workflows}d). The first execution node checks which packages are available in the battery-modeling environment. The second calls the PyBaMM solver tool and computes the discharge for the specified SPM, SPMe, or DFN model, C-rate, duration, and parameter set (for example, Chen2020\cite{chen2020lgm50}), writing a results file, a report, and a voltage curve. A guard node checks that the initial and final voltages and the voltage extrema are finite and that there are at least two sampling points. A final execution node summarizes the voltage results and the locations of the output files. The \texttt{battery-electrochem-modeler} role mounts the same solver tool and constrains the choice of model and the source of parameters through its knowledge of macroscale battery modeling and the BPX parameter specification.

\section{Background and Infrastructure for Materials Research}
\label{app:materials-background}
\subsection{Materials Research and Computational Paradigms}
Materials research studies the relationships among composition, processing, structure, and properties. Theoretical modeling and atomistic simulation allow researchers to explain microscopic mechanisms and predict behavior, while high-throughput computation extends this ability to systematic candidate screening\cite{hafner2000atomicscale,curtarolo2013hight}. Integrated computational materials engineering emphasizes coordination between materials models, manufacturing, and engineering design. The Materials Genome Initiative promotes the combination of theory, computation, synthesis, and characterization\cite{nrc2008icme,depablo2014mgi}. These developments require data and methods to be reused across stages: computational inputs, software environments, material structures, experimental conditions, and the evidence behind conclusions must travel with a task so that it can be reviewed and improved later.

\subsection{Computational Tools, Scientific Models, and Materials Data}
Table~\ref{tab:ecosystem} lists representative components of materials research infrastructure. Solvers provide physical computations, workflow frameworks handle execution and data management, and materials databases support retrieval, screening, and model training\cite{jain2013materialsproject,kirklin2015oqmd}. Scientific models extend this infrastructure. MatterGen generates inorganic crystal candidates conditioned on target properties\cite{zeni2025mattergen}. Generative models propose candidates, while machine-learning potentials predict energies and forces. Their outputs require computational or experimental validation.

\begin{table}[htbp]
\centering
\caption{Functional layers and representative systems in materials research infrastructure.}
\label{tab:ecosystem}
\begin{tabular}{p{3.6cm}p{10.5cm}}
\toprule
\textbf{Functional layer} & \textbf{Representative systems and tools} \\
\midrule
Electronic-structure computation & VASP, Quantum ESPRESSO\cite{kresse1996vasp,giannozzi2009qe} \\
Computational integration and workflows & pyiron, FireWorks, AiiDA, atomate2\cite{janssen2019pyiron,jain2015fireworks,huber2020aiida,ganose2025atomate2} \\
Open materials data & Materials Project, OQMD\cite{jain2013materialsproject,kirklin2015oqmd} \\
Structure generation & MatterGen\cite{zeni2025mattergen} \\
Input and result processing & VASPKIT\cite{wang2021vaspkit} \\
Experimental automation & A-Lab, CRESt\cite{szymanski2023alab,zhang2025crest} \\
\bottomrule
\end{tabular}
\end{table}

\subsection{Computational and Experimental Feedback Loops}
Closed loops between computation and experiment use experimental results to update the next computational or experimental decision. A-Lab combines synthesis knowledge extracted from publications with robotic solid-state synthesis and active learning\cite{szymanski2023alab}. CRESt combines composition, literature, and micrograph information to guide electrocatalyst synthesis and testing through knowledge-assisted Bayesian optimization and uses a vision-language model to diagnose experimental anomalies\cite{zhang2025crest}. These systems demonstrate closed-loop optimization in particular materials workflows. A 2026 correction to the A-Lab study, concerning phase identification and materials novelty, shows that autonomous execution records still need independent scientific verification\cite{szymanski2026alabcorrection}. In MatOS, Backend nodes submit computational and instrument jobs through the same interface (Section~\ref{sec:stateful-workflow}), which is where experimental stages would connect to a computational workflow.

\section{Supplementary Case: Reproducing the CFDBench Lid-Driven Cavity}
\label{app:cfdbench-cavity}
This appendix reports the eighth research case, a reproduction study in continuum fluid dynamics. MatOS reproduced the lid-driven cavity problem of CFDBench\cite{luo2023cfdbench}, a benchmark for machine-learning methods in fluid dynamics whose reference fields were computed with ANSYS Fluent and interpolated to a $64\times64$ grid. The cavity has well-defined geometry and boundary conditions and a classical high-resolution solution by Ghia \emph{et al.}\cite{ghia1982cavity}, so the reproduction can be checked against both the benchmark data and an independent reference.

MatOS selected three square-cavity cases with Reynolds numbers of 100, 400, and 1000 (CFDBench cases \texttt{prop/case0054}, \texttt{prop/case0060}, and \texttt{bc/case0000}). Incompressible flows with the same geometry and boundary conditions are similar at equal $\mathrm{Re}=\rho u_{\mathrm{lid}}w/\mu$, so each case was solved in nondimensional form with unit lid speed and cavity width and $\nu=1/\mathrm{Re}$. The calculations used the transient laminar solver icoFoam of OpenFOAM v2412\cite{weller1998openfoam} with PISO pressure--velocity coupling\cite{issa1986piso}, second-order central differencing for convection, and uniform $160\times160$ ($\mathrm{Re}=100$ and 400) or $224\times224$ ($\mathrm{Re}=1000$) meshes, run on 16--24 MPI processes. A run was considered steady when the initial momentum residual had reached a plateau near $10^{-11}$ and the maximum velocity change between two successive output times was at most $1.3\times10^{-7}$. Each case took 2--7~min of wall-clock time, and a serial check agreed with the parallel result to within $2.5\times10^{-7}$ in velocity.

\begin{figure}[!htb]
    \centering
    \includegraphics[width=\linewidth]{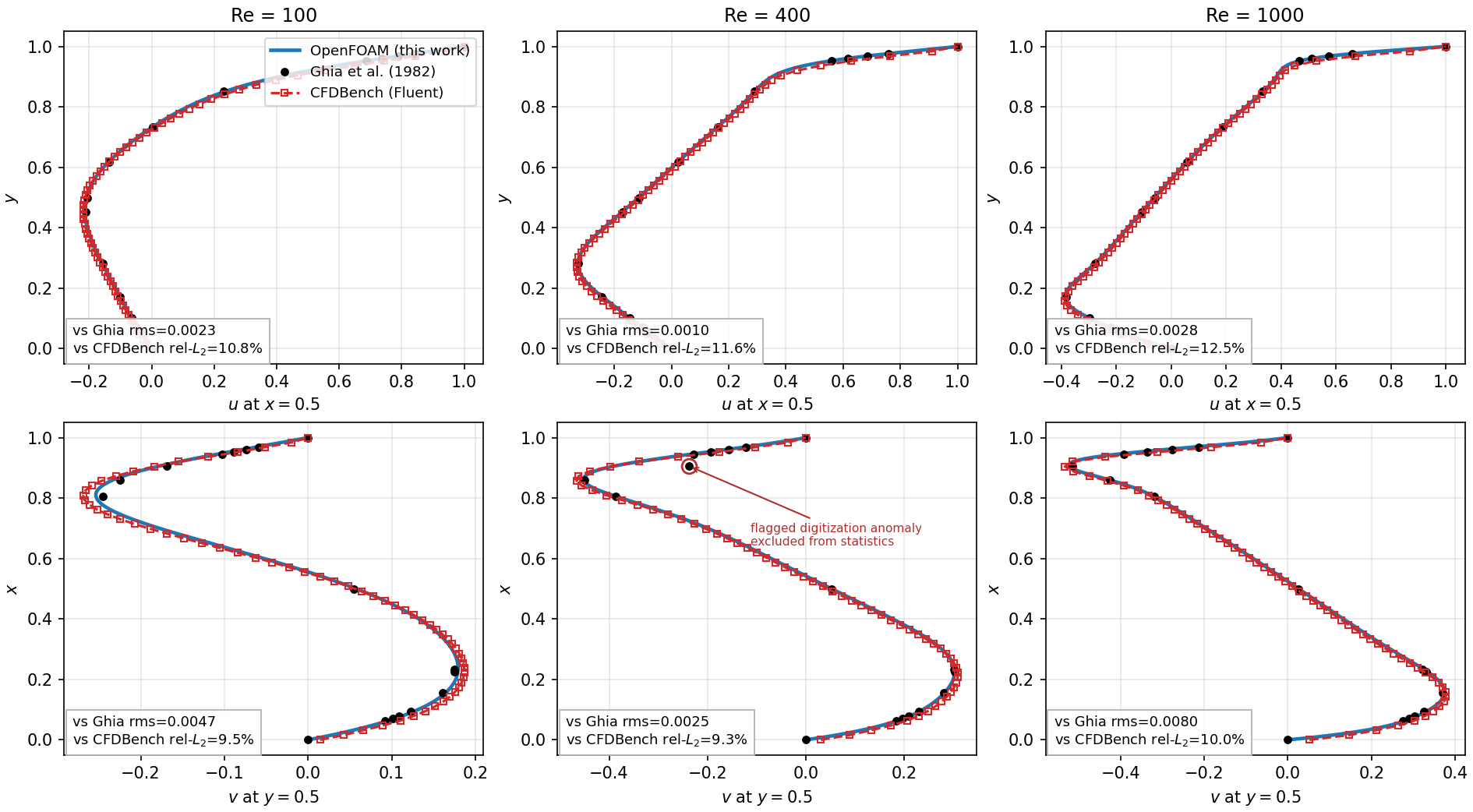}
    \small
    \caption{Centerline velocity profiles of the lid-driven cavity at $\mathrm{Re}=100$, 400, and 1000. Top panels show $u$ along $x=0.5$, and bottom panels show $v$ along $y=0.5$. Solid lines denote the OpenFOAM solution, filled circles the Ghia \emph{et al.} reference, and open squares the CFDBench data. The annotation in each panel gives the RMS deviation from Ghia \emph{et al.} and the relative $L_2$ difference from CFDBench along that centerline. These values differ from the interior-grid statistics in the text. The circled point in the $\mathrm{Re}=400$ $v$ profile is the tabulated value excluded from the statistics.}
    \label{fig:cfd-profiles}
\end{figure}

Figure~\ref{fig:cfd-profiles} compares the centerline velocity profiles with Ghia \emph{et al.} and with the stored CFDBench fields, and Table~\ref{tab:cfd-ghia} summarizes the agreement with Ghia \emph{et al.} With velocities normalized by the lid speed, the RMS deviation is 0.0010--0.0028 for $u$ along $x=0.5$ and 0.0025--0.0080 for $v$ along $y=0.5$. Primary-vortex centers agree to within 0.003 and stream-function minima to within 0.0005. One tabulated value, $v=-0.23827$ at $x=0.9063$ for $\mathrm{Re}=400$, departs from the trend of its neighbors. The same value appears in two digitizations of the table that share a common source, whereas the CFDBench solution ($-0.398$) and the present solution ($-0.388$) agree with each other. We treat the point as a suspected transcription error and exclude it from the statistics. Including it would raise the RMS deviation of the $\mathrm{Re}=400$ $v$ profile from 0.0025 to 0.0365.

\begin{table}[!htb]
    \centering
    \footnotesize
    \renewcommand{\arraystretch}{1.2}
    \caption{Agreement with Ghia \emph{et al.} for the three square-cavity cases. Velocities are normalized by the lid speed. Paired entries give the present result followed by the reference value. $^{a}$Excluding the tabulated value at $x=0.9063$ discussed in the text.}
    \label{tab:cfd-ghia}
    \begin{tabular}{@{}>{\raggedright\arraybackslash}p{5.0cm}>{\centering\arraybackslash}p{3.0cm}>{\centering\arraybackslash}p{3.0cm}>{\centering\arraybackslash}p{3.0cm}@{}}
        \toprule
        Quantity & $\mathrm{Re}=100$ & $\mathrm{Re}=400$ & $\mathrm{Re}=1000$ \\
        \midrule
        RMS deviation of $u$ along $x=0.5$ & 0.0023 & 0.0010 & 0.0028 \\
        RMS deviation of $v$ along $y=0.5$ & 0.0047 & 0.0025$^{a}$ & 0.0080 \\
        $u_{\min}$ & $-0.2138$ / $-0.2109$ & $-0.3272$ / $-0.3273$ & $-0.3864$ / $-0.3829$ \\
        $y$ at $u_{\min}$ & 0.457 / 0.453 & 0.281 / 0.281 & 0.171 / 0.172 \\
        Primary-vortex center, $x$ & 0.616 / 0.617 & 0.553 / 0.555 & 0.529 / 0.531 \\
        Primary-vortex center, $y$ & 0.734 / 0.734 & 0.603 / 0.606 & 0.565 / 0.563 \\
        $\psi_{\min}$ & $-0.1035$ / $-0.1034$ & $-0.1136$ / $-0.1139$ & $-0.1184$ / $-0.1179$ \\
        \bottomrule
    \end{tabular}
\end{table}

On the $64\times64$ benchmark grid, excluding two layers of boundary nodes, the relative $L_2$ difference between the present and stored fields ($u$ and $v$ combined) is 8.9\%, 8.1\%, and 7.6\% for $\mathrm{Re}=100$, 400, and 1000. Pointwise differences of 0.50 occur only at the two lid-corner nodes, where the two datasets assign different values to the discontinuous boundary condition. Interpolated to the sampling points of Ghia \emph{et al.}, the stored CFDBench fields themselves have RMS deviations in $u$ of 0.030, 0.041, and 0.049, an order of magnitude larger than those of the present solution. The remaining difference from the benchmark data therefore stems mainly from the interpolation and discretization error of the stored fields.

\begin{figure}[!htb]
    \centering
    \includegraphics[width=\linewidth]{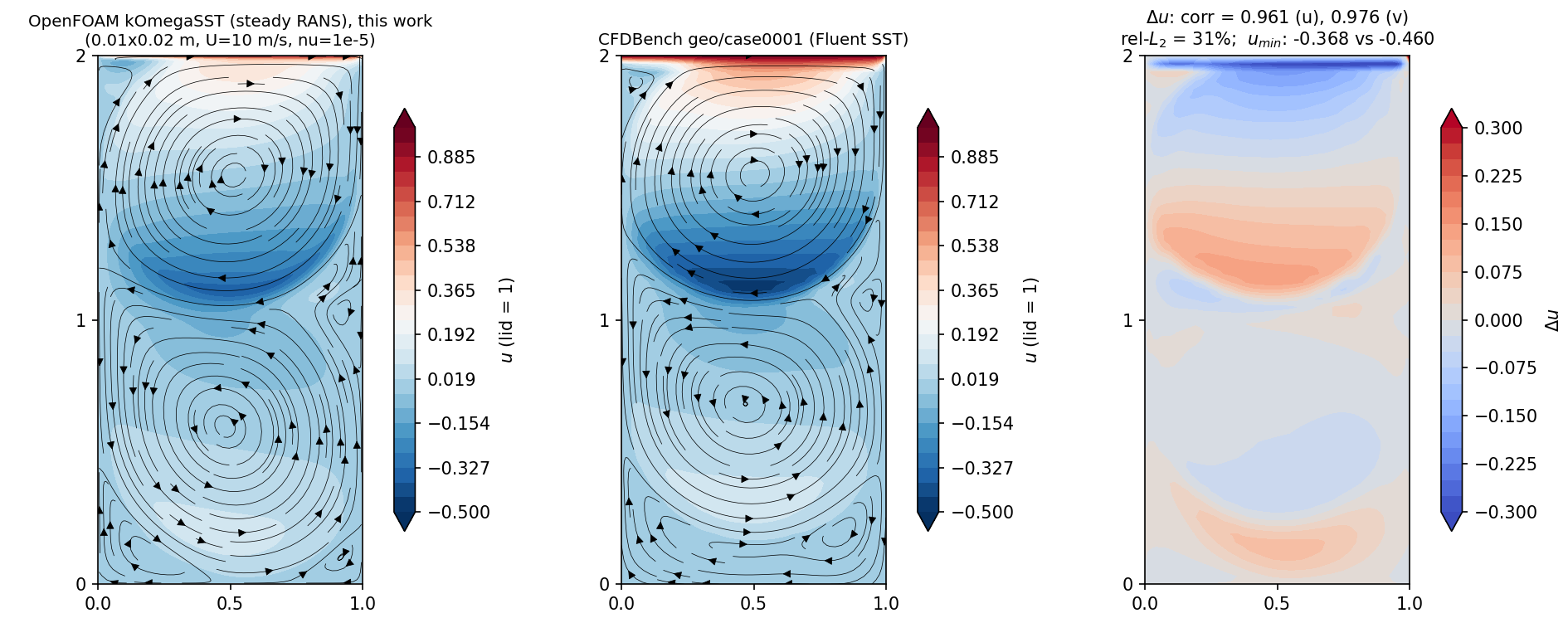}
    \small
    \caption{Steady RANS check of the rectangular cavity \texttt{geo/case0001} (width 0.01~m, height 0.02~m, $u_{\mathrm{lid}}=10$~m\,s$^{-1}$, $\nu=10^{-5}$~m$^2$\,s$^{-1}$). Left: $u$ field and streamlines from OpenFOAM with the $k$--$\omega$ SST model. Center: stored CFDBench field. Right: their difference. The right panel title gives the structure correlations, the relative $L_2$ difference (rounded), and the minimum $u$ of the two fields.}
    \label{fig:cfd-rect}
\end{figure}

While attempting a rectangular case from the geometry subset (\texttt{geo/case0001}, height-to-width ratio 2:1), MatOS found that the metadata of this subset contradict its stored fields. The case files specify a lid speed of 1~m\,s$^{-1}$, but the stored lid speed is 10~m\,s$^{-1}$ in all 24 geometry cases. The square control case \texttt{geo/case0005} has $u_{\min}=-0.4465$ near $y\approx0.03$, a profile typical of $\mathrm{Re}\sim10^4$ (Ghia \emph{et al.} report $-0.427$ near $y\approx0.06$ for $\mathrm{Re}=10{,}000$) and incompatible with laminar flow at low Reynolds number. Laminar reproductions of \texttt{geo/case0001} at $\mathrm{Re}=10$, 100, and 200 reached correlations of only 0.78--0.87 with the stored $u$ field. With the base fluid properties defined in the CFDBench paper ($u_{\mathrm{lid}}=10$~m\,s$^{-1}$, $\nu=10^{-5}$~m$^2$\,s$^{-1}$), this case corresponds to $\mathrm{Re}=10^4$. A steady RANS calculation with simpleFoam, SIMPLE pressure--velocity coupling\cite{patankar1972simple}, and the $k$--$\omega$ SST model\cite{menter1994sst} on a $128\times256$ mesh reproduces the multi-vortex structure of the stored field (Figure~\ref{fig:cfd-rect}), with correlations of 0.961 for $u$ and 0.976 for $v$ and vortex-center positions within 0.02. The remaining relative $L_2$ difference of 30.5\% is attributed to differences in turbulence modeling, near-wall treatment, and meshes. The lid-speed, density, and viscosity entries of the geometry-subset metadata should therefore be corrected before the subset is used for training or evaluation.

\section{Model Versions and Evaluation Dates}
\label{app:model-versions}
Table~\ref{tab:model-versions} lists the models, software versions, settings, and dates (UTC) of the runs reported in Section~\ref{sec:benchmarks}. GPT models were called through an OpenAI-compatible API gateway and DeepSeek models through the DeepSeek API. For GPT-5.4 and GPT-5.1, the version is the dated snapshot returned by the API. For the DeepSeek models and GPT-5.5, it is the latest release published before the run\cite{deepseek2026changelog,openai2026gpt55}. The MatOS asset library was frozen and self-evolution was disabled in all runs.

\begin{table}[!htb]
\centering
\small
\setlength{\tabcolsep}{4pt}
\renewcommand{\arraystretch}{1.15}
\caption{Models, versions, settings, and dates of the benchmark runs. Dates are in UTC.}
\label{tab:model-versions}
\begin{tabular}{@{}p{3.1cm}>{\raggedright\arraybackslash}p{2.4cm}>{\raggedright\arraybackslash}p{4.6cm}>{\raggedright\arraybackslash}p{2.6cm}p{1.6cm}@{}}
\toprule
Benchmark & System & Model and version & Setting & Dates (2026) \\
\midrule
ResearchClawBench & MatOS & GPT-5.4 (\texttt{gpt-5.4-2026-03-05}) & reasoning effort high & Sep 7--8 \\
 & MatOS & DeepSeek-V4-Flash (\texttt{DeepSeek-V4-Flash-0731}) & default & Sep 7 \\
 & Judge & GPT-5.1 (\texttt{gpt-5.1-2025-11-13}) & benchmark rubric & Sep 7--8 \\
\midrule
MatTools & MatOS & DeepSeek-V4-Pro (\texttt{DeepSeek-V4-Pro-Preview}) & 1M-token context & Jul 6 \\
 & Codex CLI 0.142.2 & DeepSeek-V4-Pro (\texttt{DeepSeek-V4-Pro-Preview}) & reasoning effort high & Jul 7 \\
 & Claude Code 2.1.191 & DeepSeek-V4-Pro (\texttt{DeepSeek-V4-Pro-Preview}) & effort max & Jul 7 \\
\midrule
SciAgentGym & MatOS & DeepSeek-V4-Pro (\texttt{DeepSeek-V4-Pro-Preview}) & 1M-token context & Jul 4--5 \\
 & Codex CLI 0.142.2 & DeepSeek-V4-Pro (\texttt{DeepSeek-V4-Pro-Preview}) & reasoning effort high & Jul 5 \\
 & Claude Code 2.1.191 & DeepSeek-V4-Pro (\texttt{DeepSeek-V4-Pro-Preview}) & effort max & Jul 6 \\
 & Judge & GPT-5.5 (\texttt{gpt-5.5-2026-04-23}) & reasoning effort extra high & Jul 5--6 \\
\bottomrule
\end{tabular}
\end{table}

On MatTools and SciAgentGym, Codex CLI reaches DeepSeek-V4-Pro through the MoonBridge adapter and Claude Code through the Anthropic-compatible DeepSeek endpoint. SciAgentGym answers are graded by six independent GPT-5.5 judges, each reviewing seven cases, with a case counted as correct only under the strict criterion.

\section{Third-Party Software, Models, and Data}
\label{app:third-party}
Tables~\ref{tab:third-party-models} and~\ref{tab:third-party-software} summarize the third-party models, datasets, software, and methods used in the benchmarks, the case studies, and Appendix~\ref{app:cfdbench-cavity}, together with their role in this report. GOL-MLP, which we developed, is listed for completeness and has no separate publication at the time of writing. The language models and agent systems used in the benchmarks are listed in Appendix~\ref{app:model-versions}.

{\footnotesize
\renewcommand{\arraystretch}{1.15}
\begin{longtable}{@{}>{\raggedright\arraybackslash}p{2.5cm}>{\raggedright\arraybackslash}p{5.2cm}>{\raggedright\arraybackslash}p{5.2cm}>{\raggedright\arraybackslash}p{1.4cm}@{}}
\caption{Models and datasets used in this report.}\label{tab:third-party-models}\\
\toprule
\textbf{Name} & \textbf{Description} & \textbf{Use in this report} & \textbf{Ref.} \\
\midrule
\endfirsthead
\caption[]{(continued)}\\
\toprule
\textbf{Name} & \textbf{Description} & \textbf{Use in this report} & \textbf{Ref.} \\
\midrule
\endhead
\midrule
\multicolumn{4}{r}{\emph{Continued on next page}}\\
\endfoot
\bottomrule
\endlastfoot
MACE & Higher-order equivariant message-passing interatomic potential & Adsorption energies in \texttt{Material\_002} and committee selection of halide configurations & \cite{batatia2022mace} \\
MatRIS-10M-OAM & Invariant, attention-based universal potential & One of the 11 evaluated potentials & \cite{zhou2026matris} \\
PET-OAM-XL & Point-edge-transformer universal potential & One of the 11 evaluated potentials & \cite{mazitov2025petmad,upet} \\
DPA-3.1-MPtrj & Line-graph neural network for large atomistic models & One of the 11 evaluated potentials & \cite{zhang2026dpa3} \\
MatterSim & Universal potential across elements, temperatures, and pressures & MD pre-screening of solid-electrolyte candidates & \cite{yang2024mattersim} \\
DeePMD-kit & Deep-potential training and MD package & Trajectory review of solid-electrolyte candidates & \cite{zeng2023deepmd} \\
ORB-v3 & Universal potential & Calculator for halide property screening & \cite{rhodes2025orbv3} \\
Allegro & Strictly local equivariant interatomic potential & Potential trained to compare MPtrj training-data selection methods & \cite{musaelian2023allegro} \\
L-G-DCNN & Composition-only model combining LSTM, GRU, and deep convolutional layers & Composition screen of solid electrolytes and alloy melting-temperature prediction & \cite{du2024lgdcnn} \\
ExtraTrees & Ensemble of extremely randomized decision trees & Alloy melting-temperature prediction baseline & \cite{geurts2006extratrees} \\
Roost & Composition-based deep representation model & Alloy melting-temperature prediction baseline & \cite{goodall2020roost} \\
CrabNet & Composition-based attention network & Alloy melting-temperature prediction baseline & \cite{wang2021crabnet} \\
Finder & Formula-graph self-attention network & Alloy melting-temperature prediction baseline, evaluated here using a continuous-fraction formula-graph implementation & \cite{ihalage2022finder} \\
DenseGNN & Deep graph neural network for crystal properties & Stability, band-gap, and elastic predictions & \cite{du2024densegnn} \\
GOL-MLP & Polarizable machine-learned force field for organic liquids with a graph-neural-network backbone, developed by the authors & Liquid-electrolyte dynamics & -- \\
MPtrj & Materials Project relaxation trajectories & Candidate structures for fixed-budget training-data selection & \cite{deng2023chgnet} \\
DIRECT & Dimensionality-reduced stratified sampling & Fixed 100,000-structure selection baseline & \cite{qi2024direct} \\
Alexandria & Open database of DFT-computed inorganic structures & Source of solid-electrolyte candidates & \cite{schmidt2024alexandria} \\
CFDBench & Benchmark of four fluid-dynamics problems with reference fields computed in ANSYS Fluent & Reference data for the lid-driven cavity & \cite{luo2023cfdbench} \\
Ghia \emph{et al.} (1982) & High-resolution multigrid solutions of the lid-driven cavity & Independent validation of centerline profiles & \cite{ghia1982cavity} \\
\end{longtable}
}

\begin{table}[htbp]
\centering
\footnotesize
\renewcommand{\arraystretch}{1.15}
\caption{Software and methods used in this report.}
\label{tab:third-party-software}
\begin{tabular}{@{}>{\raggedright\arraybackslash}p{2.5cm}>{\raggedright\arraybackslash}p{5.2cm}>{\raggedright\arraybackslash}p{5.2cm}>{\raggedright\arraybackslash}p{1.4cm}@{}}
\toprule
\textbf{Name} & \textbf{Description} & \textbf{Use in this report} & \textbf{Ref.} \\
\midrule
pymatgen & Python library for materials analysis & MatTools tasks & \cite{ong2013pymatgen} \\
pymatgen-analysis-defects & Point-defect extension of pymatgen & MatTools defect and insertion-site tasks & \cite{shen2024pmgdefects} \\
ASE & Atomic Simulation Environment & Structure construction and calculator interface & \cite{larsen2017ase} \\
RDKit (ETKDGv3) & Cheminformatics toolkit and its distance-geometry conformer generator & Conformers in the rigidity-index task & \cite{rdkit,wang2020etkdgv3} \\
SOAP & Smooth-overlap-of-atomic-positions descriptor & Environment distances and configuration novelty & \cite{bartok2013soap} \\
softBV (BVSE) & Bond-valence site-energy analysis of ion migration & Migration-barrier screening of halides & \cite{chen2019softbv} \\
NEB & Nudged-elastic-band method for minimum-energy paths & Migration barriers of halides & \cite{henkelman2000neb} \\
Packmol & Builder of initial configurations for liquid simulations & Electrolyte simulation boxes & \cite{martinez2009packmol} \\
OpenMM & GPU-accelerated MD engine & Liquid-electrolyte dynamics & \cite{eastman2024openmm} \\
ORCA & Quantum-chemistry package & Molecular descriptors of electrolyte components & \cite{neese2020orca} \\
OpenFOAM (icoFoam, simpleFoam) & Finite-volume CFD toolbox with transient laminar (PISO) and steady (SIMPLE) incompressible solvers & Lid-driven cavity reproduction & \cite{weller1998openfoam,issa1986piso,patankar1972simple} \\
$k$--$\omega$ SST & Two-equation eddy-viscosity turbulence model & Steady RANS check of the rectangular cavity & \cite{menter1994sst} \\
Pettifor scale, mat2vec & Element chemical scale and materials-text word embeddings & Novelty check in the alloy search & \cite{pettifor1984scale,tshitoyan2019unsupervised} \\
\bottomrule
\end{tabular}
\end{table}

\end{document}